\documentclass[a4paper,fleqn]{cas-sc}

\usepackage[numbers]{natbib}
\usepackage{multirow}
\usepackage{subfig}

\def\tsc#1{\csdef{#1}{\textsc{\lowercase{#1}}\xspace}}
\tsc{WGM}
\tsc{QE}
\tsc{EP}
\tsc{PMS}
\tsc{BEC}
\tsc{DE}
\begin{document}
\let\WriteBookmarks\relax
\def\floatpagepagefraction{1}
\def\textpagefraction{.001}

% Short title
\shorttitle{Prediction of morphological heart age}

% Short author
\shortauthors{Johan \"{O}fverstedt, Elin Lundstr\"{o}m, H{\aa}kan Ahlstr\"{o}m, Joel Kullberg}

% Main title of the paper
\title [mode = title]{Towards prediction of morphological heart age from computed tomography angiography}                      
% Title footnote mark
% eg: \tnotemark[1]
%\tnotemark[1,2]

% Title footnote 1.
% eg: \tnotetext[1]{Title footnote text}
% \tnotetext[<tnote number>]{<tnote text>} 
%\tnotetext[1]{This document is the results of the research
%   project funded by the National Science Foundation.}

%\tnotetext[2]{The second title footnote which is a longer text matter
%   to fill through the whole text width and overflow into
%   another line in the footnotes area of the first page.}

% First author
%
% Options: Use if required
% eg: \author[1,3]{Author Name}[type=editor,
%       style=chinese,
%       auid=000,
%       bioid=1,
%       prefix=Sir,
%       orcid=0000-0000-0000-0000,
%       facebook=<facebook id>,
%       twitter=<twitter id>,
%       linkedin=<linkedin id>,
%       gplus=<gplus id>]
\author[1]{Johan \"{O}fverstedt}[%[type=editor,
                        %auid=000,bioid=1,
                        %prefix=Sir,
                        %role=Researcher,
                        orcid=0000-0003-0253-9037]

% Corresponding author indication
\cormark[1]

% Footnote of the first author
%\fnmark[1]

% Email id of the first author
\ead{johan.ofverstedt@uu.se}

% URL of the first author
%\ead[url]{www.cvr.cc, cvr@sayahna.org}

%  Credit authorship
\credit{Conceptualization of this study, Methodology, Software, Validation, Formal analysis, Investigation, Data curation, Writing - original draft, Visualization}

% Address/affiliation
\affiliation[1]{organization={Radiological Image Analysis, Department of Surgical Sciences, Uppsala University},
    %addressline={Akademiska sjukhuset, ing{\aa}ng 70}, 
    city={Uppsala},
    % citysep={}, % Uncomment if no comma needed between city and postcode
    %postcode={751 85}, 
    % state={},
    country={Sweden}}

% Second author
\author[1]{Elin Lundstr\"{o}m}[
orcid=0000-0003-2955-4958
]

\credit{Conceptualization of this study, Validation, Writing - review and editing, Project administration, Funding acquisition}

% Third author
\author[1,2]{H{\aa}kan Ahlstr\"{o}m}[%
orcid=0000-0002-8701-969X
   ]

\credit{Conceptualization of this study, Validation, Resources, Writing - review and editing, Supervision, Project administration, Funding acquisition}

\author[1,2]{Joel Kullberg}[
orcid=0000-0001-8205-7569
]
\credit{Conceptualization of this study, Validation, Resources, Writing - review and editing, Supervision, Project administration, Funding acquisition}

% Address/affiliation
\affiliation[2]{organization={Antaros Medical},
    % addressline={}, 
    city={M\"{o}lndal},
    % citysep={}, % Uncomment if no comma needed between city and postcode
    %postcode={695014}, 
    %state={Trivandrum},
    country={Sweden}}

% Corresponding author text
\cortext[cor1]{Corresponding author}
%\cortext[cor2]{Principal corresponding author}

% Footnote text
%\fntext[fn1]{This is the first author footnote. but is common to third
%  author as well.}
%\fntext[fn2]{Another author footnote, this is a very long footnote and
%  it should be a really long footnote. But this footnote is not yet
%  sufficiently long enough to make two lines of footnote text.}

% For a title note without a number/mark
%\nonumnote{This note has no numbers. In this work we demonstrate $a_b$
%  the formation Y\_1 of a new type of polariton on the interface
%  between a cuprous oxide slab and a polystyrene micro-sphere placed
%  on the slab.
%  }

% Here goes the abstract
\begin{abstract}
Age prediction from medical images or other health-related non-imaging data is an important approach to data-driven aging research, providing knowledge of how much information a specific tissue or organ carries about the chronological age of the individual. In this work, we studied the prediction of age from computed tomography angiography (CTA) images, which provide detailed representations of the heart morphology, with the goals of (i) studying the relationship between morphology and aging, and (ii) developing a novel \emph{morphological heart age} biomarker.

We applied an image registration-based method that standardizes the images from the whole cohort into a single space. We then extracted supervoxels (using unsupervised segmentation), and corresponding robust features of density and local volume, which provide a detailed representation of the heart morphology while being robust to registration errors. Machine learning models are then trained to fit regression models from these features to the chronological age.

We applied the method to a subset of the images from the Swedish CArdioPulomonary bioImage Study (SCAPIS) dataset, consisting of 721 females and 666 males. We observe a mean absolute error in comparison to the chronological age of $2.74$ years for females and $2.77$ years for males. The predictions from different sub-regions of interest were observed to be more highly correlated with the predictions from the whole heart, compared to the chronological age, revealing a high consistency in the predictions from morphology. 

Saliency analysis was also performed on the prediction models to study what regions are associated positively and negatively with the predicted age. This resulted in detailed association maps where the density and volume of known, as well as some novel sub-regions of interest, are determined to be important. The saliency analysis aids in the interpretability of the models and their predictions.
\end{abstract}

% Use if graphical abstract is present
% \begin{graphicalabstract}
% \includegraphics{figs/grabs.pdf}
% \end{graphicalabstract}

% Research highlights
%\begin{highlights}
%\item Research highlights item 1
%\item Research highlights item 2
%\item Research highlights item 3
%\end{highlights}

% Keywords
% Each keyword is seperated by \sep
\begin{keywords}
age prediction \sep regression \sep supervoxels \sep image registration \sep computed tomography angiography
%quadrupole exciton \sep polariton \sep \WGM \sep \BEC
\end{keywords}

\maketitle
\section{Introduction}

Aging research is advancing at many levels of granularity from the (sub-)cellular to morphological and functional aspects of tissue and organs. One prominent approach to data-driven aging research is age prediction, and a number of works in recent years have proposed various methodologies for predicting the age of an individual from a variety of organs (or anatomical regions of interest) and input modalities such as magnetic resonance imaging (MRI) \cite{langner2019identifying,ecker2024deep,sihag2024explainable,starck2023atlas}, computed tomography (CT) \cite{kerber2023deep,pickhardt2024novel}, electrocardiography (ECG) \cite{lima2021deep}.

A common approach shared by many machine learning-based age prediction methodologies is to fit a model from image data (or other features) to the chronological age of the individual at the time of the image acquisition. These models tend to be developed for two main purposes, where they first can aid the assessment of how much information about the individual's chronological age is present in the input data, and through saliency analysis answer questions about what aspects of the data is most important for such predictions \cite{langner2019identifying}. A second purpose for these models is to investigate if they, once trained to predict chronological age, are able to capture clinically relevant aspects of a more health-related biological age, and thus be a superior predictor of future health outcomes compared to the chronological age alone \cite{lima2021deep,sihag2024explainable}.

\begin{figure}[t]
\centering
\includegraphics[width=\linewidth]{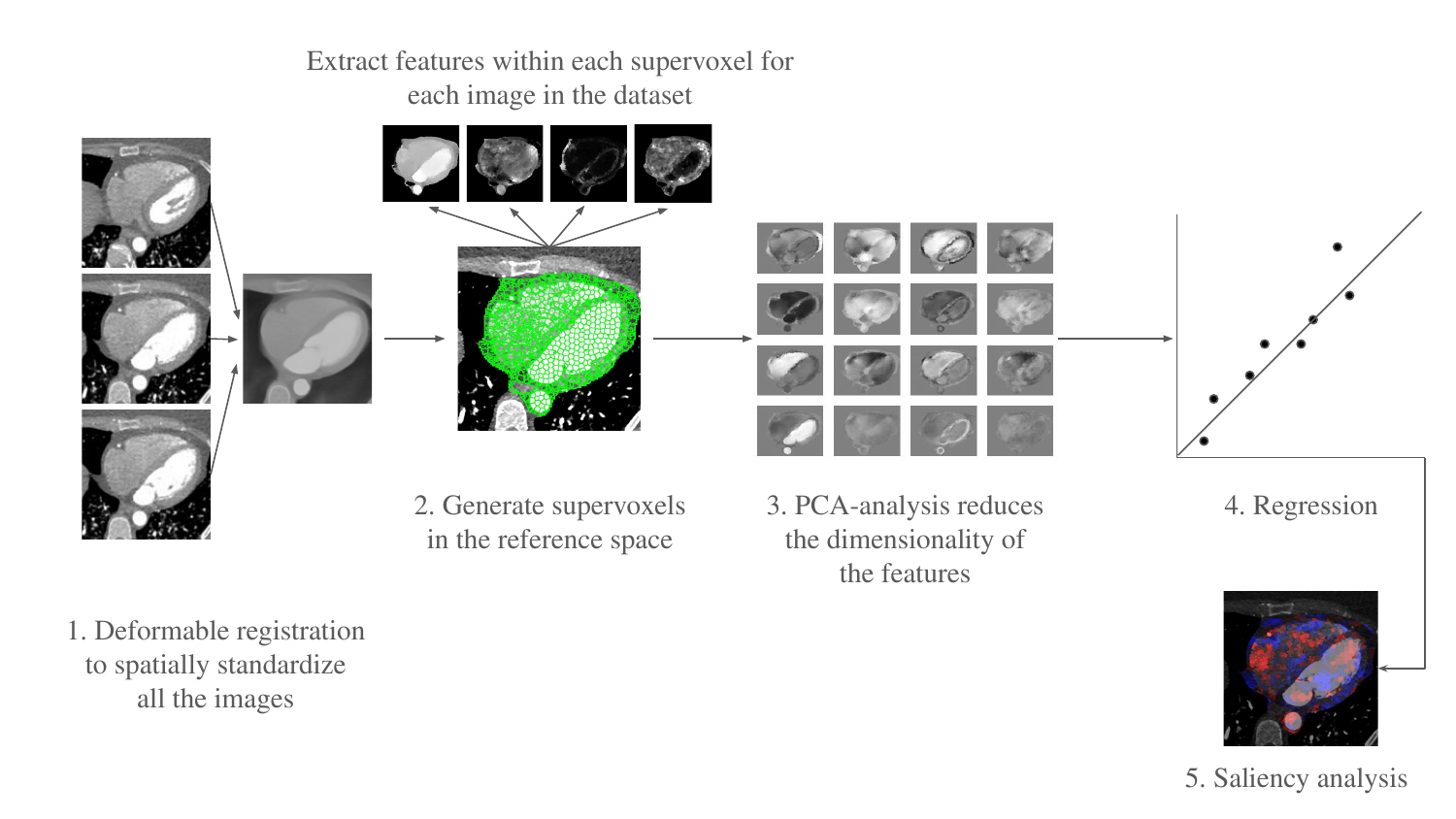}
\caption{Illustration of the main components of the analysis pipeline. Step 1: The images are registered to a common template space. Step 2: A supervoxel segmentation mask is generated, followed by feature extraction of robust statistics (median and robust standard deviation) from each supervoxel. Step 3: Principal component analysis (PCA) is applied to the supervoxel-derived features. Step 4: Linear regression is applied to these spatially standardized low-dimensional features to predict the target value. Step 5: Saliency analysis is performed by mapping the saliency of each feature back into image space to visualize the strength (and sign) of the association between each feature and the target. The entire analysis is applied to males and females separately.}
\label{fig:main}
\end{figure}

Computed tomography angiography (CTA) is an imaging methodology that generates detailed sub-${\mathtt{mm}}^3$ resolution 3D image data of the heart's morphology (chambers, myocardium, aorta, and coronary vessels) through contrast-medium enhanced CT images \cite{min2010present}.

In this work, we aim to
\begin{itemize}
    \item Develop a machine learning method for age prediction from detailed morphological CTA image data, by using compact image representations through spatially standardized supervoxel features.
    \item Perform a proof-of-concept study to investigate the performance of imaging target (explicit volume measurement) prediction for testing the ability of the feature extraction and machine learning methodology to retrieve known heart morphology information in the input features.
    \item Evaluate how well chronological age can be predicted from the morphological image features of the \emph{whole heart}, or selected sub-regions, and measure the level of agreement between predictions made from different sub-regions.
    \item Through saliency analysis, determine which features contribute to the predictions (positively and negatively), and visualize these relationships.
\end{itemize}

% In the manuscript, the female slice is 192, and the male slice is 112.

\section{Methods and materials}
Figure \ref{fig:main} illustrates the major parts of the developed analysis pipeline used in this work.

\subsection{Data}

SCAPIS is a large cohort study of a random sample ($n=30154$) of men and women aged 50-64 years in Sweden \cite{bergstrom2015swedish}.

For this study, we used a subset of the full cohort ($n=1387$, $n_{\mathtt{FEMALE}}=721$, $n_{\mathtt{MALE}}=666$) who were examined with CTA at Uppsala University Hospital, Uppsala, Sweden. 

For each subject, we had access to one 3D CTA image volume of axial slices (size: $512 \times 512$, number of slices: between 350 and 550), slice spacing approximately $0.30\texttt{ mm}$, and a slice thickness typically $0.33\texttt{ mm}$ (with minor variation across the dataset), as well as their age in whole months at the time of the examination and their (legal) sex. The CTA imaging protocol included the injection of an intravenous contrast media before the image acquisition. The time delay between injection and start of the acquisition was determined from a test bolus to match the start time with the contrast arrival at the ascending aorta, to standardize the contrast distribution between subjects.

\subsubsection{Ethics}

Ethics approval was obtained from the Swedish Ethical Review Authority (Dnr 2022-07308-01) to conduct this research study related to human subjects, with associated sex and age information. All subjects provided informed written consent for their collected data to be used for research and for that research to be published. The study adheres to the Declaration of Helsinki. SCAPIS has been approved as a multicentre trial by the ethics committee at Ume{\aa} University and adheres to the Declaration of Helsinki \cite{bergstrom2015swedish}.

\subsection{Image segmentation and inter-subject registration}

First, each CTA volume was segmented \cite{chen2020deep} using the general-purpose tool TotalSegmentator v1 \cite{wasserthal2023totalsegmentator}, which delineates a body into 104 different anatomical regions. For this work, we mostly focus on TotalSegmentator's ability to segment the heart into the four chambers: left ventricle (LV), left atrium (LA), right ventricle (RV), and right atrium (RA), as well as delineate the myocardium (MYO) and aorta (both ascending and descending combined).

For each sex, a representative template image near the center of the distributions in terms of size of various volumetric properties was selected, and the image volumes were registered to a common template space using a deformable registration method described in detail in \cite{ofverstedt2024method}, based on graph cuts \cite{tang2007non,ekstrom2020fast,ekstrom2021faster,jonsson2022image}, including a combination of the CTA images as well as masks generated using TotalSegmentator. The registration was found to have very good robustness and overall performance, with very few failed registrations \cite{ofverstedt2024method}.

\subsection{Spatially standardized image representations based on supervoxels}

Following the idea in \cite{ofverstedt2024method} of using supervoxels extracted from the CTAs, we extend the methodology from supervoxel-wise analysis to whole heart regression of target parameters, incorporating all the spatially standardized supervoxels in a single regression process.

To extract supervoxels from the reference images, we used the \emph{Simple Linear Iterative Clustering} (SLIC) method \cite{achanta2012slic}, as implemented in \texttt{SimpleITK} \cite{beare2018image}. We set the spatial proximity factor to $0.2$, with voxel-wise transformation of the CT image values $I(x, y, z)$ as
\begin{equation}
    \hat{I}(x, y, z) = \frac{\min\big(200, \max\big(-300, I(x, y, z)\big)\big)}{300}.
\end{equation}
The SLIC grid size parameter defines how densely the initial cluster centroids are placed, and therefore sets an upper limit on how many supervoxels the image can be subdivided into. We found through an ablation study that a grid size parameter of 14, yielding $14990$ supervoxels for females and $14553$ supervoxels for males, gave good overall downstream regression performance in early tests (as evaluated on prediction of chronological age), and is the default choice in the rest of the study. 

\begin{figure}[ht]
    \centering
    \subfloat[][Supervoxels]{
    \includegraphics[width=0.18\linewidth]{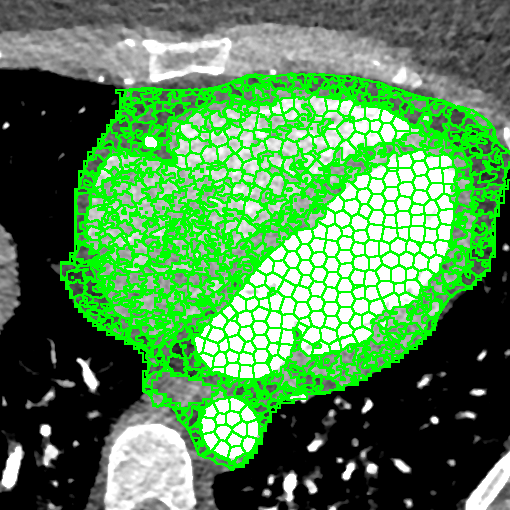}} \hfill
    \subfloat[][Median density (HU)]{
    \includegraphics[width=0.18\linewidth]{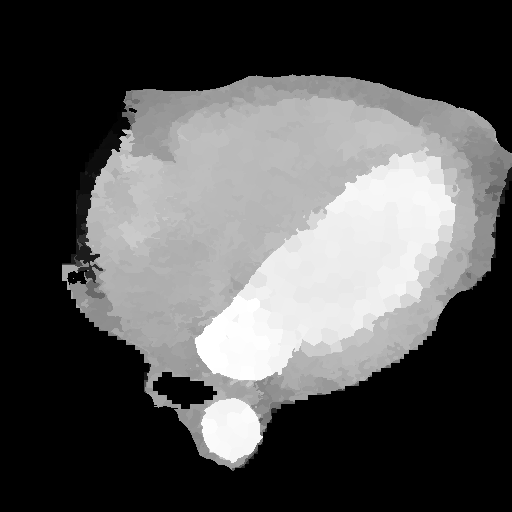}} \hfill
    \subfloat[][Median local volume (JacDet)]{
    \includegraphics[width=0.18\linewidth]{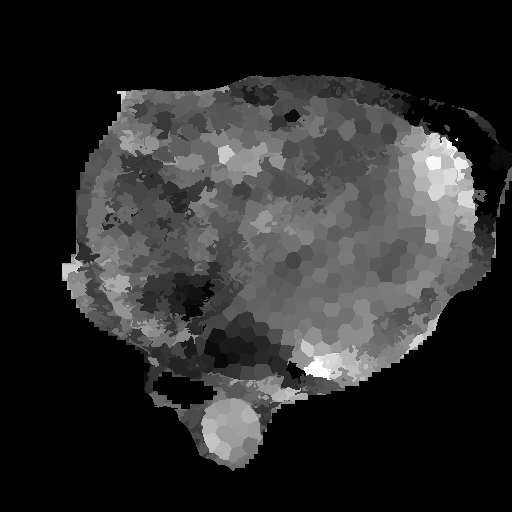}} \hfill
    \subfloat[][Stddev density (HU)]{
    \includegraphics[width=0.18\linewidth]{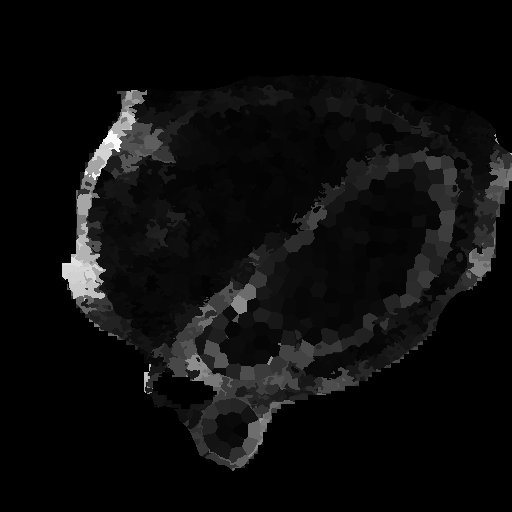}} \hfill
    \subfloat[][Stddev local volume (JacDet)]{
    \includegraphics[width=0.18\linewidth]{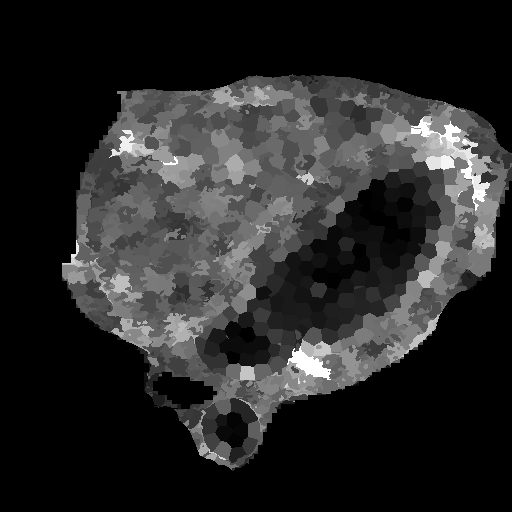}}
    \caption{Illustration of the supervoxels delineated over an axial slice of the female template (a), and features (b,c,d,e) for a randomly selected female subject, mapped to and displayed in the template space.}
    \label{fig:features}
\end{figure}

For each supervoxel, we extracted four features: median density, median Jacobian determinant (JacDet), which describes local volume expansion/compression of each template voxel induced by the deformation field, a robust standard deviation of density, a robust standard deviation of JacDet. The robust standard deviation is here defined as the standard deviation of the remaining values within a supervoxel after performing 1.5 IQR filtering of the density values, aiming in particular to avoid computing statistics over voxels from different tissue types in a single supervoxel. For brevity, we denote these features as standard deviation (or Stddev) throughout this work, even though it refers to the robust standard deviation. The supervoxels and the four features are illustrated in Fig.~\ref{fig:features}.

\subsection{Regions of interest}

We investigated different regions of interest (ROIs) within the hearts for two main reasons: (i) to study how well the chronological age can be predicted from only the features derived from each sub-region, and (ii) to study the agreement of age predictions from the different sub-regions. Table \ref{tab:rois} describes the different ROIs.

The ROIs were defined based on the TotalSegmentator segmentations of the template images. The surrounding non-heart tissue like lungs, stomach, esophagus, vertebrae, sternum, ribs, and non-cardiac adipose tissue were segmented using a combination of TotalSegmentator and manual delineation in the template images and used to filter out those regions such that features were only derived based on heart morphology. The region \emph{other} consists of all voxels that are in the heart but not in any of the other segmentation regions (LV, RV, LA, RA, MYO, Aorta) and is composed of various tissues or structures such as pericardial fat, and coronary vessels. Supervoxels comprised of fewer than 50 voxels (after filtering) were removed to avoid the inclusion of noisy features located on the boundary of the regions.

\begin{table}[ht]
    \centering
    \caption{A table of all sub-ROIs of the heart that we use in this work. We give a name to each, which is used to refer to the region in later sections.}
    \label{tab:rois}
    %\resizebox{0.65\linewidth}{!}{
    \begin{tabular}{c|c|c|c|c|c|c|c}
        ROI name & LV & RV & LA & RA & MYO & Aorta & Other \\ \hline \hline
        \multirow{2}{*}{Whole Heart} & \multirow{2}{*}{$\checkmark$} & \multirow{2}{*}{$\checkmark$} & \multirow{2}{*}{$\checkmark$} & \multirow{2}{*}{$\checkmark$} & \multirow{2}{*}{$\checkmark$} & \multirow{2}{*}{$\checkmark$}  & \multirow{2}{*}{$\checkmark$} \\
        & & & & & & & \\ \hline
        LV, RV, LA, RA, & \multirow{2}{*}{$\checkmark$} & \multirow{2}{*}{$\checkmark$} & \multirow{2}{*}{$\checkmark$} & \multirow{2}{*}{$\checkmark$} & \multirow{2}{*}{$\checkmark$} & \multirow{2}{*}{$\checkmark$}  & \multirow{2}{*}{-} \\
        MYO, Aorta & & & & & & & \\ \hline
        \multirow{2}{*}{LV, RV, LA, RA} & \multirow{2}{*}{\checkmark} & \multirow{2}{*}{\checkmark} & \multirow{2}{*}{\checkmark} & \multirow{2}{*}{\checkmark} & \multirow{2}{*}{-} & \multirow{2}{*}{-}  & \multirow{2}{*}{-} \\
        & & & & & & & \\ \hline
        \multirow{2}{*}{Only Other} & \multirow{2}{*}{-} & \multirow{2}{*}{-} & \multirow{2}{*}{-} & \multirow{2}{*}{-} & \multirow{2}{*}{-} & \multirow{2}{*}{-}  & \multirow{2}{*}{$\checkmark$} \\  
         & & & & & & & \\ \hline
        \multirow{2}{*}{LV} & \multirow{2}{*}{\checkmark} & \multirow{2}{*}{-} & \multirow{2}{*}{-} & \multirow{2}{*}{-} & \multirow{2}{*}{-} & \multirow{2}{*}{-}  & \multirow{2}{*}{-} \\
        & & & & & & & \\ \hline
        \multirow{2}{*}{RV} & \multirow{2}{*}{-} & \multirow{2}{*}{\checkmark} & \multirow{2}{*}{-} & \multirow{2}{*}{-} & \multirow{2}{*}{-} & \multirow{2}{*}{-}  & \multirow{2}{*}{-} \\
        & & & & & & & \\ \hline
        \multirow{2}{*}{LA} & \multirow{2}{*}{-} & \multirow{2}{*}{-} & \multirow{2}{*}{\checkmark} & \multirow{2}{*}{-} & \multirow{2}{*}{-} & \multirow{2}{*}{-}  & \multirow{2}{*}{-} \\
        & & & & & & & \\ \hline
        \multirow{2}{*}{RA} & \multirow{2}{*}{-} & \multirow{2}{*}{-} & \multirow{2}{*}{-} & \multirow{2}{*}{\checkmark} & \multirow{2}{*}{-} & \multirow{2}{*}{-}  & \multirow{2}{*}{-} \\
        & & & & & & & \\ \hline
        \multirow{2}{*}{MYO} & \multirow{2}{*}{-} & \multirow{2}{*}{-} & \multirow{2}{*}{-} & \multirow{2}{*}{-} & \multirow{2}{*}{\checkmark} & \multirow{2}{*}{-}  & \multirow{2}{*}{-} \\
        & & & & & & & \\ \hline
        \multirow{2}{*}{Aorta} & \multirow{2}{*}{-} & \multirow{2}{*}{-} & \multirow{2}{*}{-} & \multirow{2}{*}{-} & \multirow{2}{*}{-} & \multirow{2}{*}{\checkmark}  & \multirow{2}{*}{-} \\
        & & & & & & & \\ \hline
    \end{tabular}%}
\end{table}

\subsection{Linear regression based on explicit measurements of morphological features derived through segmentation}

A first regression experiment was performed using TotalSegmentator-derived features (segment mean density and volume) from six sub-structures of the heart: LV, RV, LA, RA, MYO, and the aorta. We trained a linear model from these 12 features to predict the chronological age of each individual, using \texttt{scikit-learn} \cite{scikit-learn}.

\subsection{Linear regression based on spatially standardized supervoxels}

Having computed the supervoxel-based morphological features (capturing composition and shape of the tissue) in standardized image space, we can use machine learning methodologies that are not invariant to translation and other geometrical transformations, which would not be feasible to use on non-standardized image data.

To reduce the dependency on the absolute magnitude of the features, we perform a z-score transformation on each feature, as well as on the target parameter, based on the statistics of the training set for each cross-validation fold. The same mean and standard deviation values derived from the training set partition are then used to scale the corresponding feature and target values on the validation set. The ability to scale each individual feature across the training set is an advantage made possible by spatial standardization, where each supervoxel represents the same spatial region (up to registration errors).

To attenuate the influence of very small or very large feature values in the dataset caused by noise or registration errors which can distort the features of individual supervoxels, we perform a hard clipping of each z-score at $\pm 1 \sigma$ for the age prediction task. The optimal level at which to clip the z-scores will vary depending on the prediction target. For targets that are expected to have a high linear correlation with at least a subset of the features, such as volume measurements of various anatomical substructures, we applied clipping at $\pm 3 \sigma$, restricting the influence of outliers while avoiding the removal of critical information. 

Given a clip level $c$, and the aggregate statistics of feature $x_i$ of $\mu_{x_i}$ and empirical standard deviation $\sigma_{x_i}$, we transform the features according to 
\begin{equation}
z_i({x_i}; c) = \max(-c\sigma_{x_i}, \min(c\sigma_{x_i}, \frac{x-\mu_{x_i}}{\sigma_{x_i}})).
\end{equation}

\begin{figure}[ht]
    \centering
    \includegraphics[width=0.9\linewidth]{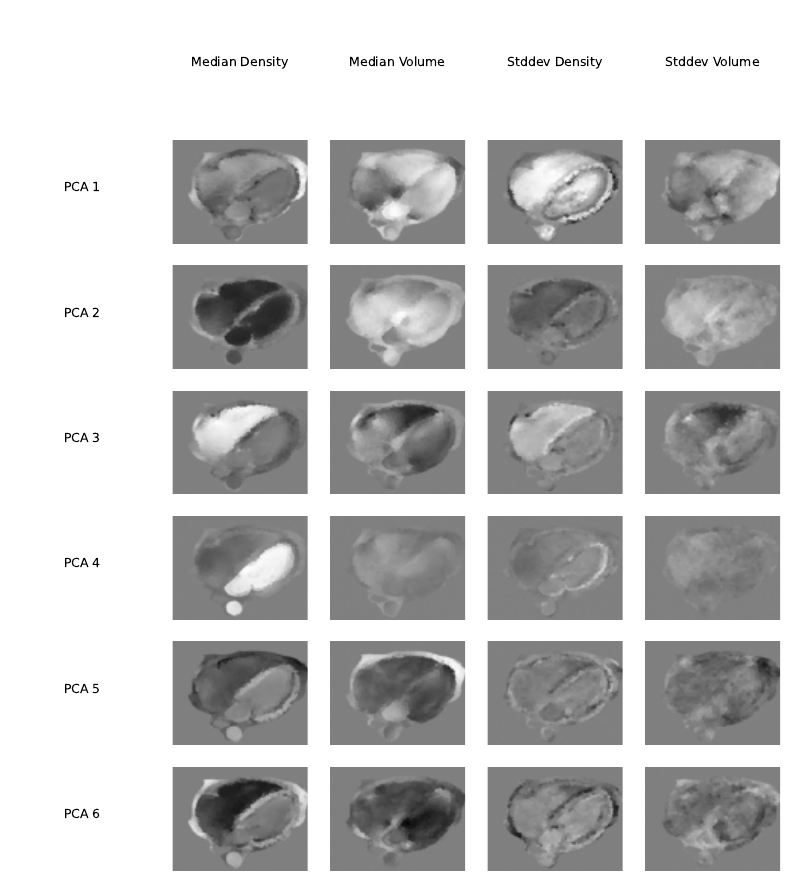}
    \caption{Illustration of the first 6 PCA components for the Supervoxel-PCA representations of the female sub-group. Gray (as seen in the background) represents no contribution to the PCA component, black a negative contribution, and white a positive contribution.}
    \label{fig:pca}
\end{figure}

To reduce the dimensionality of the feature vectors, to both (i) reduce overfitting and (ii) reduce the noise of the saliency maps, we used a methodology inspired by Eigen faces \cite{sirovich1987low} where a new smaller set of basis-vectors for the images is derived through unsupervised learning. We used Principal Component Analysis (PCA) \cite{scikit-learn} (using the implementation in \texttt{scikit-learn} called \texttt{IncrementalPCA}) for reducing the number of features to a user-selected number of components in a fast and stable manner. The PCA dimensionality reduction is applied to the whole set of four features per supervoxel, meaning that principal components may correspond to combinations of multiple features and supervoxels. An illustration of the PCA applied to the supervoxels is presented in \ref{fig:pca}. For experiments based on all or most of the supervoxels in the heart, we use 550 components, and for the experiments where we look at a single region such as LV or aorta, we use only 64 components, to avoid including components containing little useful information. The exact number of components to use is mostly an empirical question, and we arrived at the chosen values through a few preliminary tests. In the appendix, an ablation study is included, investigating the impact of this choice. 

We used an unregularized (least-squares) linear regression model from the PCA representations of the images to fit the chronological age \cite{scikit-learn}, enabled by spatial standardization, which gives a fixed role and meaning to each feature across the cohort. In Appendix D, we also compare the linear models to multi-layer neural networks, with several activation functions (non-linearities).

In addition to the age prediction, we also performed a proof-of-concept study, where we predicted six target variables describing the measured volumes LVV, RVV, LAV, RAV, MYOV, and AV, to evaluate how well the feature extraction and regression method works, given that this information is derived from the images and should be possible to predict from a set of features that preserve the important image information. The same 25-fold cross-validation scheme is used for the proof-of-concept study as for the age prediction study.

\clearpage

\subsection{Evaluation measures and reference values}

We evaluated the regression performance using several measures: the mean absolute error (MAE) between predicted values and reference values, the coefficient-of-determination $R^2$, and the Spearman correlation coefficient between predictions and reference values.

To evaluate the age prediction, we use the chronological age as the reference value (and for model fitting). To evaluate the volume prediction, we use the TotalSegmentator-derived volume measurements as ground truth.

\subsection{Cross-validation}

We conducted the proof-of-concept and age prediction studies through 25-fold cross-validation. We randomly split the dataset into 25 equally sized folds separately for females and males. PCA components and normalization/clipping were all performed on the training partitions. It is left as future work on a larger dataset to validate the methodology on additional unseen data. 25 folds were used as a trade-off between computational resources and the effective size of the training partitions, which consist of most samples already at 25 folds with diminishing returns.

\subsection{System specifications and software versions}

All experiments were performed on a single machine with Intel i9-13900K, 16 cores, 5800 MHz maximum clock frequency, 64 GB RAM, Nvidia A4000 GPU, running Ubuntu 22.04. The versions of the main software libraries are Numpy 1.25.2, Scipy 1.11.4, SimpleITK 2.3.1, pandas 2.1.3, Scikit-image 0.22.0, Scikit-learn 1.3.2, TotalSegmentator 1.5.7, and pydeform 0.5.1 (req. SimpleITK 2.3.0). 

\subsection{Ablation experiments}

We performed a number of ablation experiments to understand which parameter settings and design choices are critical to the performance of the predictions. They include:
\begin{itemize}
\item Feature subsets - The selection of subsets of the supervoxel features to use. We considered the subsets \{All features, Median density, Median volume, Stddev density, Stddev volume, Median density and volume, Stddev density and volume, Median density and Stddev density, Median volume and Stddev volume\}.%- correlations presented in Appendix A
\item Grid size - The number of supervoxels (grid-size). We considered grid sizes $\{10, 11, \ldots, 24, 25\}$.
\item Clipping level - We considered values $c\in\{0.25, 0.5, \ldots 4.75, 5.0\}$.
\item The number of PCA components. We considered values \\$n_{\mathtt{PCA}}\in\{50, 100, \ldots 550, 600\}$.
%\item Age range
\end{itemize}

\subsection{Saliency analysis}

We performed saliency analysis by mapping the regression coefficients back to the supervoxels (in image space), through the PCA transformation, such that we could visualize the (signed) importance of each feature overlayed on the template images.
The saliency maps are computed from the regression coefficients $\beta_{\texttt{\tiny{PCA}},i}$, and the PCA weights $w_i(x)$, denoting the weight of feature $x$ in PCA component $i$, as
\begin{equation}
    \hat{S}(x) = \sum\limits_{i=1}^{N_\texttt{\tiny{PCA}}}w_i(x)\beta_{\texttt{\tiny{PCA}},i},
\end{equation}
followed by normalization through division by the absolute value of the largest such saliency,
\begin{equation}
    S(x) = \frac{\hat{S}(x)}{\max\limits_{y} |\hat{S}(y)|}.
\end{equation}
Finally, these saliency maps are aggregated by computing the average saliency of each feature over all the cross-validation folds, to obtain a single saliency map for the entire dataset, where each saliency value is in the range $[-1, +1]$.

\section{Results}

\subsection{Linear regression based on explicit measurements}

In Tab.~\ref{tab:regressionexplicit}, we present the performance obtained from applying linear regression to explicit measurements of mean density (HU) and volume (mL) from segmented regions (LV, RV, LA, RA, MYO, Aorta). The $\mathtt{MAE}$ is $3.28$ and $3.34$ (years) for females and males, respectively, and $R^2$ is $0.217$ and $0.154$ for females and males, respectively.

\begin{table}[]
    \centering
    \caption{Performance of chronological age prediction using multiple linear regression and explicit measurements of mean density and volume of salient regions (LV, RV, LA, RA, MYO, Aorta) derived using TotalSegmentator. The performance measures used are MAE (in years), $R^2$, and the Spearman correlation coefficient $\rho_s$. This experiment is carried out using cross-validation over 25 folds. $\downarrow$ denotes that a lower value is better and $\uparrow$ denotes that a higher value is better.}
    \label{tab:regressionexplicit}
    %\resizebox{\linewidth}{!}{
    \begin{tabular}{c|c|c|c|c|c}
         Sex & $n$ & ROI & $\texttt{MAE} \downarrow$ & $R^2 \uparrow$ & $\rho_s \uparrow$ \\\hline \hline
         Female & $721$ & \multirow{2}{*}{LV, RV, LA, RA, MYO, Aorta} & $3.28$ & $0.217$ & $0.473$ \\
         Male & $666$ &  & $3.34$ & $0.154$ & $0.399$ \\
    \end{tabular}%}
\end{table}

\subsection{Linear regression based on spatially standardized supervoxels}

First, we present the proof-of-concept study results in Tab.~\ref{tab:poc_volume}. All the targets exhibit excellent performance with $R^2$ between $0.911$ (for the aorta volume) and $0.987$ (for the myocardium volume), with the lower performance of aorta volume prediction likely due to its position near the border of the field of view, leading to variability in both its measurement and the aorta features.

\begin{table}[]
    \centering
    \caption{Performance of proof-of-concept prediction of volumetric parameters using multiple linear regression and supervoxel-based representations corresponding to the \emph{whole heart}, extracted in a standardized geometry achieved by image registration. The performance measures used are MAE (in mL), $R^2$, and the Spearman correlation coefficient $\rho_s$. This experiment is performed using cross-validation over 25 folds. $\downarrow$ denotes that a lower value is better and $\uparrow$ denotes that a higher value is better.}
    \begin{tabular}{c|c|c|c|c|c|c}
        Sex & $n$ & ROI & Target & $\texttt{MAE} \downarrow$ & $R^2 \uparrow$ & $\rho_s \uparrow$ \\ \hline \hline
        Female & 721 & \multirow{2}{*}{Whole Heart} & \multirow{2}{*}{LVV} & 1.31 & 0.982 & 0.993 \\
        Male & 666 & & & 1.72 & 0.986 & 0.993 \\  \hline
        Female & 721 & \multirow{2}{*}{Whole Heart} & \multirow{2}{*}{RVV} & 2.90 & 0.967 & 0.983 \\
        Male & 666 & & & 3.60 & 0.976 & 0.985 \\  \hline
        Female & 721 & \multirow{2}{*}{Whole Heart} & \multirow{2}{*}{LAV} & 2.17 & 0.947 & 0.974 \\ 
        Male & 666 & & & 2.31 & 0.954 & 0.977 \\  \hline
        Female & 721 & \multirow{2}{*}{Whole Heart} & \multirow{2}{*}{RAV} & 2.70 & 0.940 & 0.971 \\ 
        Male & 666 & & & 3.00 & 0.961 & 0.981 \\  \hline
        Female & 721 & \multirow{2}{*}{Whole Heart} & \multirow{2}{*}{MYOV} & 1.40 & 0.985 & 0.993 \\ 
        Male & 666 & & & 1.86 & 0.987 & 0.993 \\  \hline
        Female & 721 & \multirow{2}{*}{Whole Heart} & \multirow{2}{*}{AV} & 3.23 & 0.915 & 0.960 \\ 
        Male & 666 & & & 5.29 & 0.911 & 0.957  \\ 
    \end{tabular}
    \label{tab:poc_volume}
\end{table}

\begin{figure}[ht]
    \centering
    \subfloat[][Chronological age (years) prediction performance (Female): $\mathtt{MAE}=2.74$, $R^2=0.436$.]{
    \includegraphics[width=0.48\linewidth]{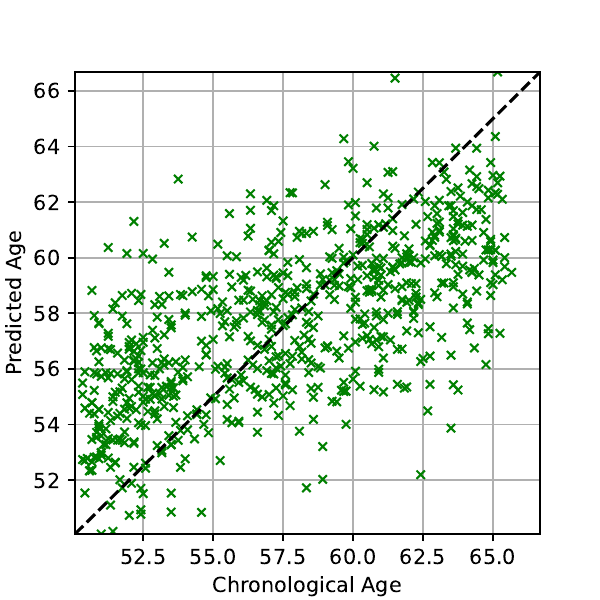}}\hfill
    \subfloat[][Chronological age (years) prediction performance (Male): $\mathtt{MAE}=2.77$, $R^2=0.381$.]{
    \includegraphics[width=0.48\linewidth]{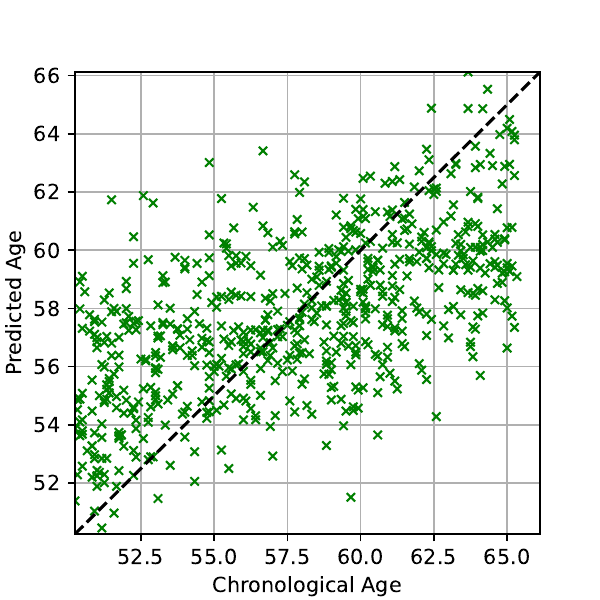}}
    \caption{Scatter plot of chronological age prediction results using the supervoxel-based features from the \emph{whole heart}.}
    \label{fig:scatter_plots}
\end{figure}

In Tab.~\ref{tab:regressionsupervoxels}, we present the performance of the supervoxel-based age prediction compared with the chronological age, predicted from the selected ROIs. The $R^2$ from the whole heart is $0.436$ for females and $0.381$ for males with lower performance for the individual regions, of which predictions from the aorta provide the highest performance, with an $R^2$ of $0.286$ for females and $0.285$ for males. The predictions from the \emph{only other} region exhibit high performance with $0.400$ for females and $0.367$ for males, which is not much lower than the predictions from the \emph{whole heart}. Figure \ref{fig:scatter_plots} displays scatter plots for the age prediction from the \emph{whole heart}. We observe an underestimated slope compared to the diagonal line.

\begin{table}[ht]
    \centering
    \caption{Performance of chronological age prediction using multiple linear regression and supervoxel-based representations of various ROI:s of the heart, extracted in a standardized geometry achieved by image registration. The performance measures used are MAE (in years), $R^2$, and the Spearman correlation coefficient $\rho_s$. This experiment is performed using cross-validation over 25 folds. $\downarrow$ denotes that a lower value is better and $\uparrow$ denotes that a higher value is better.}
    \label{tab:regressionsupervoxels}
    %\resizebox{\linewidth}{!}{
    \begin{tabular}{c|c|c|c|c|c}
        Sex & $n$ & ROI & $\texttt{MAE} \downarrow$ & $R^2 \uparrow$ & $\rho_s \uparrow$ \\ \hline \hline
        Female & 721 & \multirow{2}{*}{Whole Heart} & $\mathbf{2.74}$ & $\mathbf{0.436}$ & $\mathbf{0.666}$ \\
        Male & 666 &  & $\mathbf{2.77}$ & $\mathbf{0.381}$ & $\mathbf{0.614}$ \\ \hline \hline
        Female & 721 & \multirow{2}{*}{LV, RV, LA, RA, MYO, Aorta} & $2.78$ & $0.427$ & $0.656$ \\
        Male & 666 &  & $2.89$ & $0.350$ & $0.592$ \\ \hline
        Female & 721 & \multirow{2}{*}{LV, RV, LA, RA} & $2.99$ & $0.348$ & $0.591$ \\
        Male & 666 &  & $3.11$ & $0.239$ & $0.499$ \\ \hline 
        Female & 721 & \multirow{2}{*}{Only Other} & $2.86$ & $0.400$ & $0.636$ \\
        Male & 666 &   & $2.81$ & $0.367$ & $0.602$ \\ \hline \hline
        Female & 721 & \multirow{2}{*}{LV} & $3.27$ & $0.228$ & $0.478$ \\
        Male & 666 & & $3.43$ & $0.118$ & $0.352$ \\ \hline
        Female & 721 & \multirow{2}{*}{RV} & $3.40$ & $0.193$ & $0.435$ \\
        Male & 666 & & $3.33$ & $0.161$ & $0.401$ \\ \hline
        Female & 721 & \multirow{2}{*}{LA} & $3.30$ & $0.217$ & $0.461$ \\
        Male & 666 & & $3.40$ & $0.147$ & $0.391$ \\ \hline
        Female & 721 & \multirow{2}{*}{RA} & $3.65$ & $0.059$ & $0.259$ \\
        Male & 666 & & $3.54$ & $0.077$ & $0.285$ \\ \hline
        Female & 721 & \multirow{2}{*}{MYO} & $3.30$ & $0.213$ & $0.464$ \\
        Male & 666 &  & $3.34$ & $0.158$ & $0.397$ \\ \hline
        Female & 721 & \multirow{2}{*}{Aorta} & $3.08$ & $0.286$ & $0.534$ \\
        Male & 666 & & $3.07$ & $0.285$ & $0.530$ \\
    \end{tabular}%}
\end{table}

\begin{figure}[ht]
    \centering
    \includegraphics[width=1.0\linewidth]{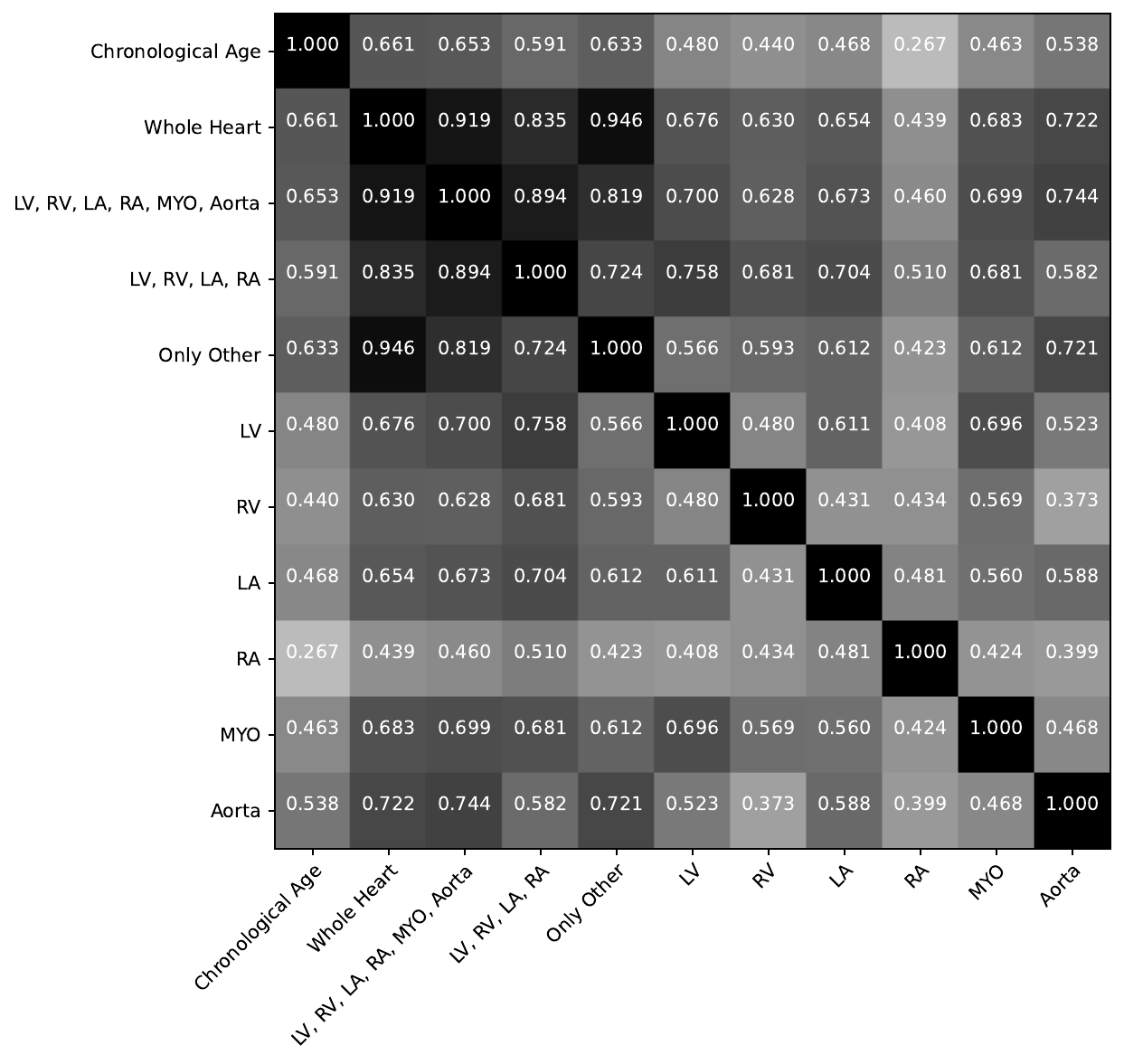}
    \caption{Pearson correlation between the age prediction of various spatial sub-ROIs, as well as chronological age for the female sub-group.}
    \label{fig:roicorrelationfemale}
\end{figure}

\begin{figure}[ht]
    \centering
    \includegraphics[width=1.0\linewidth]{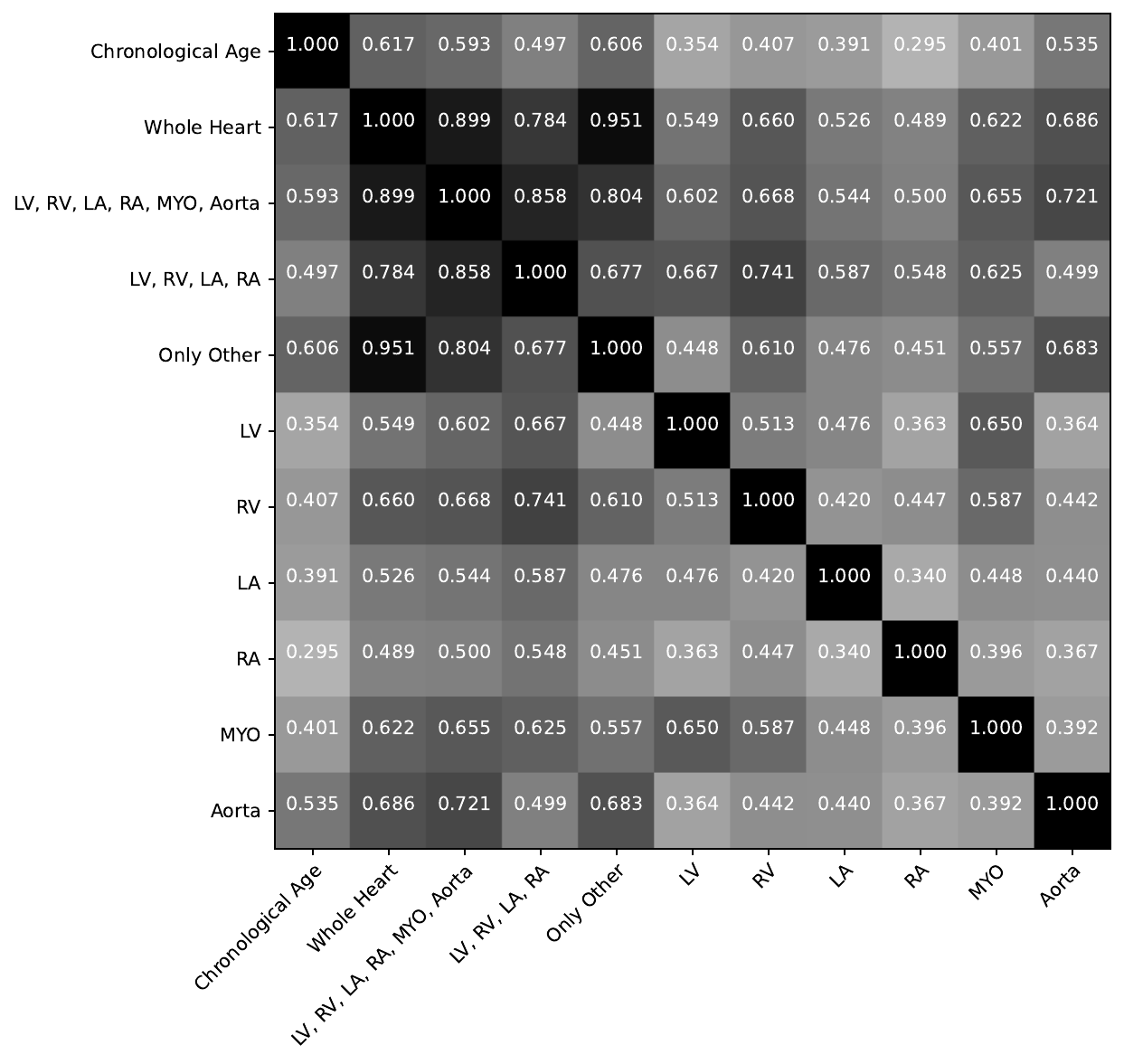}
    \caption{Pearson correlation between the age prediction of various spatial sub-ROIs, as well as chronological age for the male sub-group.}
    \label{fig:roicorrelationmale}
\end{figure}

Figures \ref{fig:roicorrelationfemale} and \ref{fig:roicorrelationmale} display the correlation between the age prediction from several ROI:s (as well as chronological age) \cite{ecker2024deep}. The correlation between the age predictions is high ($>0.89$) for the ROIs that exhibit relatively highest performance: (LV, RV, LA, RA, Myo, Aorta) and \emph{Only Other}, with lower correlation between the predictions from the \emph{whole heart} and the other individual sub-regions. The correlation coefficients between predictions from the whole heart and predictions from each sub-region are significantly higher (p-values are presented in Tab.~\ref{tab:fishertests}) than the corresponding correlation coefficients between the predictions from the sub-regions and chronological age.

\begin{table}[ht]
    \centering
    \caption{Statistical analysis of pairwise differences in correlation coefficients between the regression from the \emph{whole heart} ROI and each other selected ROI, and the chronological age and each selected ROI, corresponding to pairwise testing of the first two rows of Fig.~\ref{fig:roicorrelationfemale} and Fig.~\ref{fig:roicorrelationmale}. A two-sided Fisher z-test is used, adjusted for multiple testing (18 tests) using Bonferroni correction, yielding a P-value threshold of $0.0028$. We denote the $\rho_{\mathtt{CA}}$ as the correlation between chronological age and the predicted values corresponding to each listed ROI, and $\rho_{\mathtt{WH}}$ the correlation between the predictions from the \emph{whole heart} and the predicted values of each listed ROI. In each test, $\rho_{\mathtt{WH}}$ is statistically significantly different from $\rho_{\mathtt{CA}}$.}
    \label{tab:fishertests}
    %\resizebox{\linewidth}{!}{
    \begin{tabular}{c|c|c|c|c|c|c}
         Sex & n & $\rho_{\mathtt{CA}}$ & $\rho_{\mathtt{WH}}$ & ROI & P-value & Sign. \\ \hline \hline
         Female & 721 & 0.653 & 0.919 & \multirow{2}{*}{LV, RV, LA, RA, MYO, Aorta} & $p<0.0001$ & \checkmark \\
         Male & 666 & 0.593 & 0.899 & & $p<0.0001$ & \checkmark \\ \hline
         Female & 721 & 0.591 & 0.835 & \multirow{2}{*}{LV, RV, LA, RA} & $p<0.0001$ & \checkmark \\
         Male & 666 & 0.497 & 0.784 & & $p<0.0001$ & \checkmark \\ \hline
         Female & 721 & 0.633 & 0.946 & \multirow{2}{*}{Only Other} & $p<0.0001$ & \checkmark \\
         Male & 666 & 0.606 & 0.951 &  & $p<0.0001$ & \checkmark \\ \hline
         Female & 721 & 0.480 & 0.676 & \multirow{2}{*}{LV} & $p<0.0001$ & \checkmark \\
         Male & 666 & 0.354 & 0.549 & & $p<0.0001$ & \checkmark \\ \hline
         Female & 721 & 0.440 & 0.630 & \multirow{2}{*}{RV} & $p<0.0001$ & \checkmark \\
         Male & 666 & 0.407 & 0.660 &  & $p<0.0001$ & \checkmark \\ \hline
         Female & 721 & 0.468 & 0.654 & \multirow{2}{*}{LA} & $p<0.0001$ & \checkmark \\
         Male & 666 & 0.391 & 0.526 & & $p<0.002$ & \checkmark \\ \hline
         Female & 721 & 0.267 & 0.439 &  \multirow{2}{*}{RA} & $p<0.001$ & \checkmark \\
         Male & 666 & 0.295 & 0.489 & & $p<0.0001$ & \checkmark \\ \hline
         Female & 721 & 0.463 & 0.683 & \multirow{2}{*}{MYO} & $p<0.0001$ & \checkmark \\
         Male & 666 & 0.401 & 0.622 & & $p<0.0001$ & \checkmark \\ \hline
         Female & 721 & 0.538 & 0.722 &\multirow{2}{*}{Aorta} & $p<0.0001$ & \checkmark \\
         Male & 666 & 0.535 & 0.686 & & $p<0.0001$ & \checkmark \\ 
    \end{tabular}
    %}
\end{table}

\begin{table}[ht]
    \centering
    \caption{Ablation study of chronological age prediction using multiple linear regression and subsets of the supervoxel-based features for the \emph{whole heart} ROI. The performance measures used are MAE (in years), $R^2$, and the Spearman correlation coefficient $\rho_s$. This experiment is performed using cross-validation over 25 folds. $\downarrow$ denotes that a lower value is better and $\uparrow$ denotes that a higher value is better.}
    \label{tab:regressionsupervoxelsfeatureablation}
    %\resizebox{\linewidth}{!}{
    \begin{tabular}{c|c|c|c|c|c|c}
        Sex & $n$ & ROI &  Features & $\texttt{MAE} \downarrow$ & $R^2 \uparrow$ & $\rho_s \uparrow$ \\ \hline \hline
        Female & 721 & \multirow{2}{*}{Whole Heart} & \multirow{2}{*}{All Features} & $\mathbf{2.74}$ & $\mathbf{0.436}$ & $\mathbf{0.666}$ \\
        Male & 666 & &  & $\mathbf{2.77}$ & $\mathbf{0.381}$ & $\mathbf{0.614}$ \\ \hline \hline         
        Female & 721 & \multirow{2}{*}{Whole Heart} & \multirow{2}{*}{Median Density} & $2.88$ & $0.377$ & $0.621$ \\
        Male & 666 & &  & $3.02$ & $0.267$ & $0.540$ \\ \hline 
        Female & 721 & \multirow{2}{*}{Whole Heart} & \multirow{2}{*}{Median Volume} & $3.07$ & $0.296$ & $0.565$ \\
        Male & 666 & &  & $3.09$ & $0.237$ & $0.532$ \\ \hline

        Female & 721 & \multirow{2}{*}{Whole Heart} & \multirow{2}{*}{Stddev Density} & $2.94$ & $0.353$ & $0.596$ \\
        Male & 666 & &  & $2.94$ & $0.319$ & $0.564$ \\ \hline 
        Female & 721 & \multirow{2}{*}{Whole Heart} & \multirow{2}{*}{Stddev Volume} & $3.13$ & $0.289$ & $0.542$ \\
        Male & 666 & &  & $3.14$ & $0.225$ & $0.477$ \\ \hline \hline 

        Female & 721 & \multirow{2}{*}{Whole Heart} & \multirow{2}{*}{Median and Stddev Density} & $2.85$ & $0.392$ & $0.631$ \\
        Male & 666 & & & $2.84$ & $0.357$ & $0.601$ \\ \hline
        
        Female & 721 & \multirow{2}{*}{Whole Heart} & \multirow{2}{*}{Median and Stddev Volume} & $2.97$ & $0.352$ & $0.597$ \\
        Male & 666 & & & $2.98$ & $0.297$ & $0.545$ \\ \hline \hline

        Female & 721 & \multirow{2}{*}{Whole Heart} & \multirow{2}{*}{Median Density and Volume} & $2.83$ & $0.403$ & $0.643$ \\
        Male & 666 & & & $2.82$ & $0.349$ & $0.599$ \\ \hline
        
        Female & 721 & \multirow{2}{*}{Whole Heart} & \multirow{2}{*}{Stddev Density and Volume} & $2.85$ & $0.399$ & $0.637$ \\
        Male & 666 & & & $2.89$ & $0.343$ & $0.584$ \\ \hline \hline 

        % without chambers
        Female & 721 & \multirow{2}{*}{Only Other} & \multirow{2}{*}{Median and Stddev Density} & $2.87$ & $0.381$ & $0.622$ \\
        Male & 666 & &  & $2.81$ & $0.356$ & $0.597$ \\ \hline 
        Female & 721 & \multirow{2}{*}{Only Other} & \multirow{2}{*}{Median and Stddev Volume} & $3.12$ & $0.295$ & $0.557$ \\
        Male & 666 & &  & $3.14$ & $0.230$ & $0.493$ \\
    
    \end{tabular}
    %}
\end{table}

\subsection{Ablation studies}

In Tab.~\ref{tab:regressionsupervoxelsfeatureablation}, we present the result of the ablation study related to the choice of supervoxel features. The density features are the most important (in the sense of providing higher performance when removing the other features), with median density being the most important single feature for females and Stddev of density being the most important feature for males. Even excluding the chambers, myocardium, and aorta, the density features remain more important than the volume features.

In Appendix C, we display the results of varying the key hyper-parameters. We found that the performance was insensitive to the choice of the tested hyper-parameters, and thus no extensive fine-tuning is required.

In Appendix D, we experiment with deep learning models and compare them with the linear models used in the main part of the study. We observe that linear models perform similarly to or better than the tested deep learning models.

In Appendix E, we display the results of an additional ablation study, where we filter out the subjects who were outliers in the proof-of-concept experiment in terms of MAE. Even when we filtered out those subjects, the performance was very similar to the performance without filtering.

\clearpage

\subsection{Saliency analysis}

\begin{figure}[ht]
    \centering
    \includegraphics[width=0.8\linewidth]{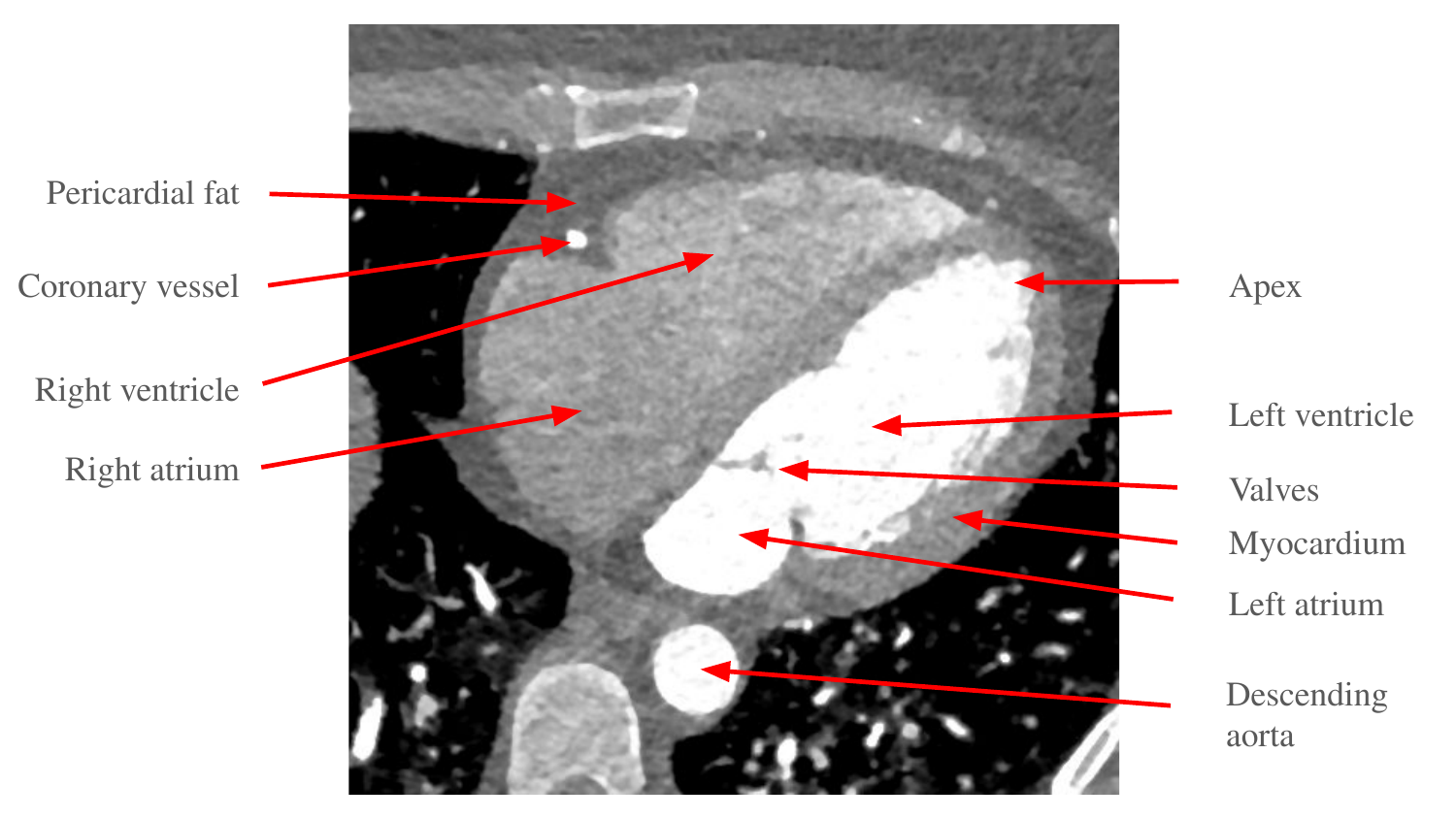}
    \caption{A diagram of the main cardiac structures/regions discussed in the saliency analysis, here shown on an axial slice for the female template.}
    \label{fig:heartanatomy}
\end{figure}

\begin{figure}[ht]
    \centering
    \textbf{Saliency maps for LVV prediction (FEMALE)}\\
    \subfloat[][Median density]{
    \includegraphics[width=0.24\linewidth]{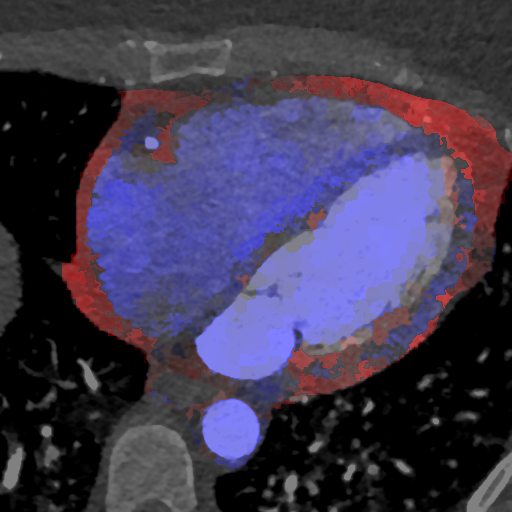}}
    \subfloat[][Median volume]{
    \includegraphics[width=0.24\linewidth]{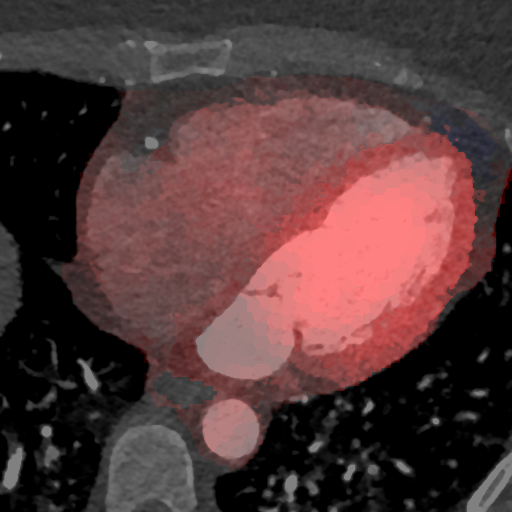}}
    \subfloat[][Stddev density]{
    \includegraphics[width=0.24\linewidth]{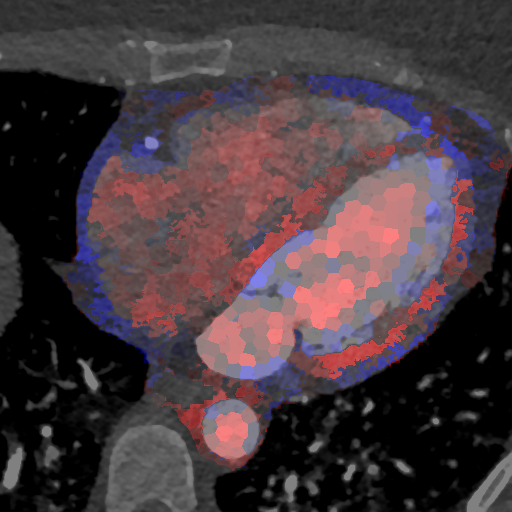}}
    \subfloat[][Stddev volume]{
    \includegraphics[width=0.24\linewidth]{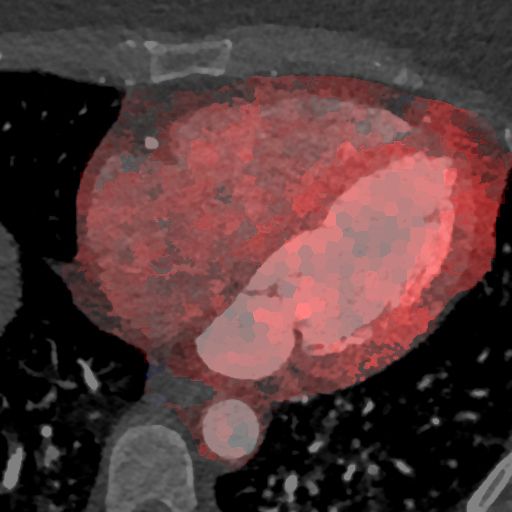}}\\\vspace{8mm}
    \textbf{Saliency maps for LVV prediction (MALE)}\\ 
    \subfloat[][Median density]{
    \includegraphics[width=0.24\linewidth]{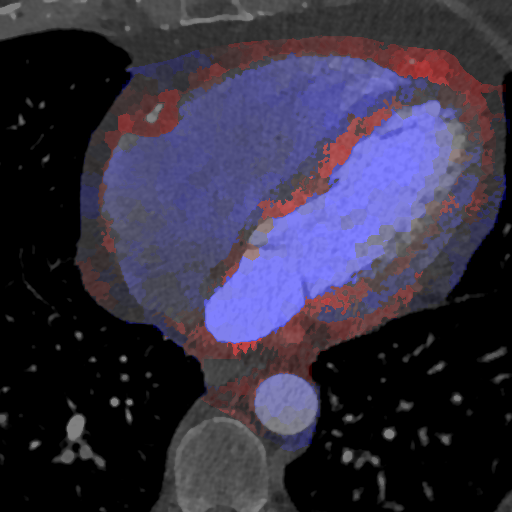}}
    \subfloat[][Median volume]{
    \includegraphics[width=0.24\linewidth]{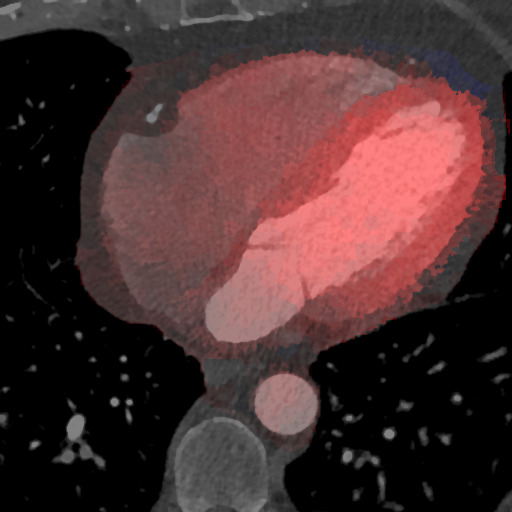}}
    \subfloat[][Stddev density]{
    \includegraphics[width=0.24\linewidth]{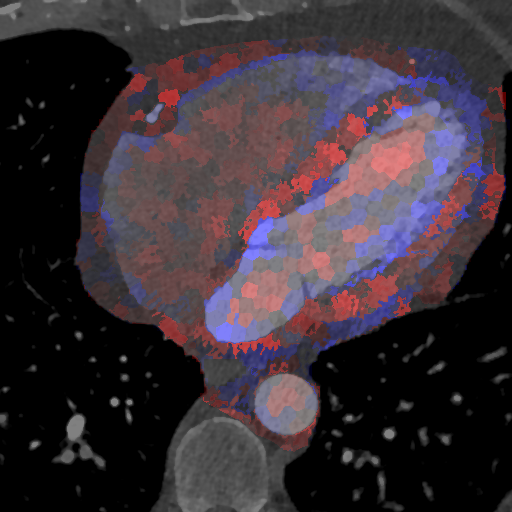}}
    \subfloat[][Stddev volume]{
    \includegraphics[width=0.24\linewidth]{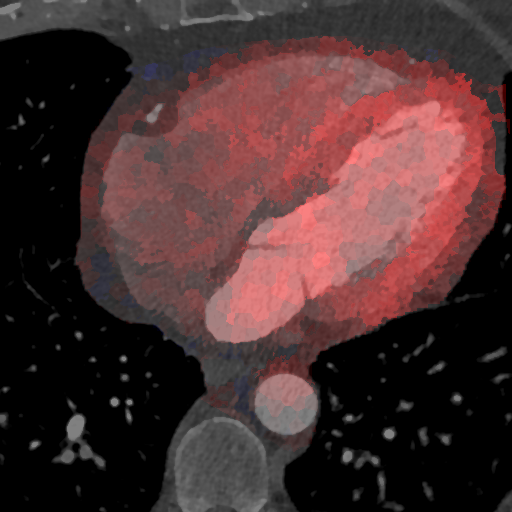}}\\
    \hfill\includegraphics[width=0.3\linewidth]{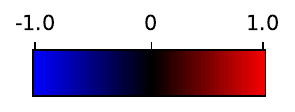}
    \caption{An axial slice of the saliency map overlayed on the CT image (windowed to -512 to +512 HU) for the four features per supervoxel used as input for the proof-of-concept prediction of LVV, with red representing a positive association and blue a negative association (colorless representing non-saliency).}% and a positive association between the volume in the myocardium and LVV.}
    \label{fig:lvv_saliency_maps}
\end{figure}

\begin{figure}[ht]
    \centering
    \textbf{Saliency maps for age prediction (FEMALE)}\\
    \subfloat[][Median density]{
    \includegraphics[width=0.24\linewidth]{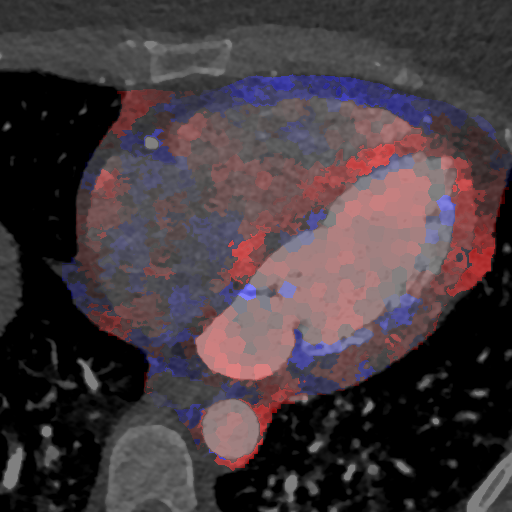}}
    \subfloat[][Median volume]{
    \includegraphics[width=0.24\linewidth]{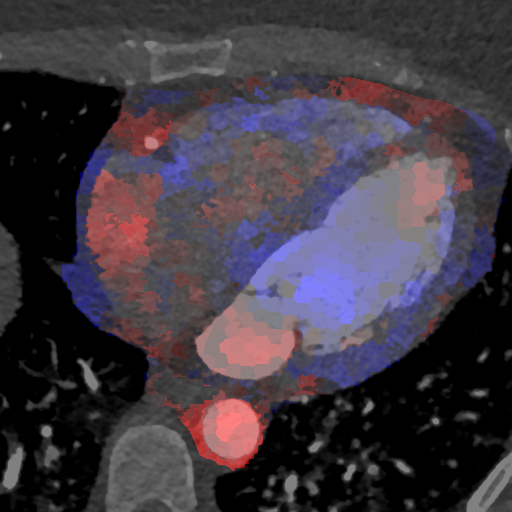}}
    \subfloat[][Stddev density]{
    \includegraphics[width=0.24\linewidth]{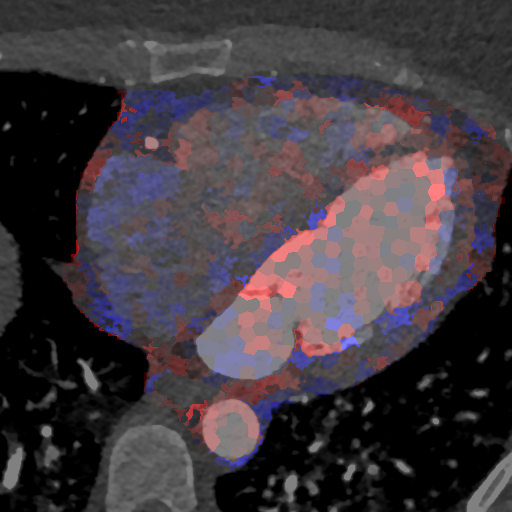}}
    \subfloat[][Stddev volume]{
    \includegraphics[width=0.24\linewidth]{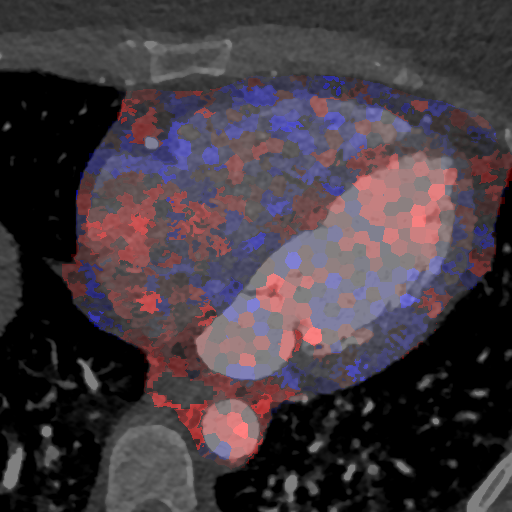}}\\ \vspace{8mm}
    \textbf{Saliency maps for age prediction (MALE)}\\
    \subfloat[][Median density]{
    \includegraphics[width=0.24\linewidth]{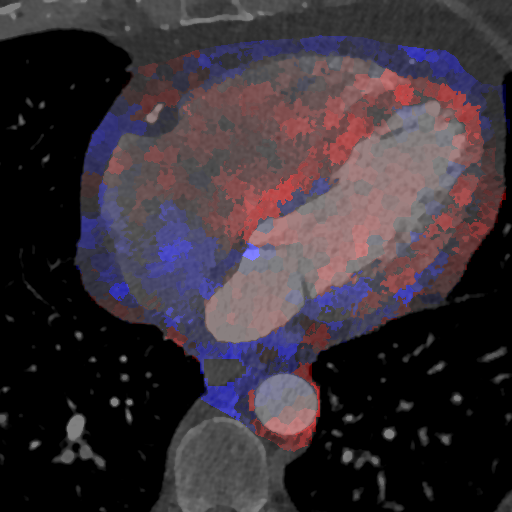}}
    \subfloat[][Median volume]{
    \includegraphics[width=0.24\linewidth]{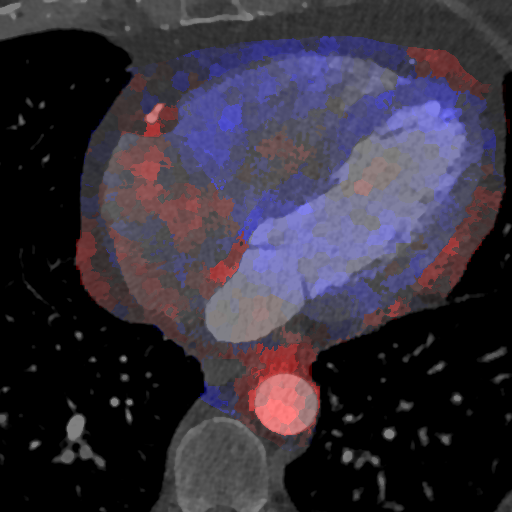}}
    \subfloat[][Stddev density]{
    \includegraphics[width=0.24\linewidth]{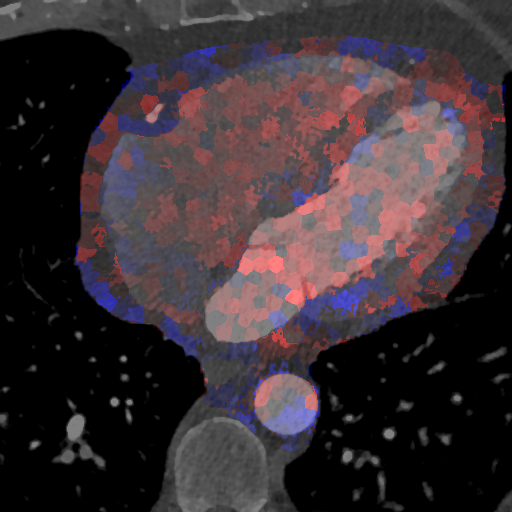}}
    \subfloat[][Stddev volume]{
    \includegraphics[width=0.24\linewidth]{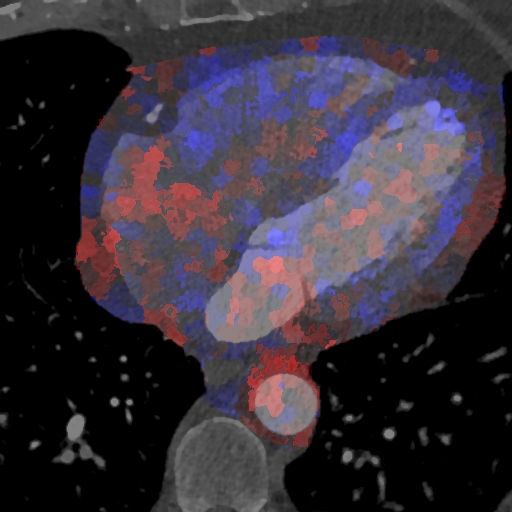}}\\
    \hfill\includegraphics[width=0.3\linewidth]{figs/saliency_colormap.pdf}
    \caption{An axial slice of the saliency map overlayed on the CT image (windowed to -512 to +512 HU) for the four features per supervoxel used as input for age prediction, with red representing a positive association and blue a negative association (colorless representing non-saliency).
    }
    \label{fig:age_saliency_maps}
\end{figure}

We present a single selected axial slice (containing LV, LA, RA, RV, MYO, and Aorta) as illustrated in Fig.~\ref{fig:heartanatomy}, of the saliency map volumes related to the four features computed for each supervoxel for the two sexes, resulting in a total of 8 saliency maps displayed in Fig.~\ref{fig:lvv_saliency_maps}, and Fig.~\ref{fig:age_saliency_maps}. Full saliency map volumes for the age prediction as well as for all the proof-of-concept experiments are available as supplementary material \cite{ofverstedt_2024_13832059}.

The main findings from the saliency analysis for LVV prediction are:
\begin{itemize}
    \item The volume features within the LV is highlighted as the most important (with a positive sign).
    \item The volume features within the myocardium is highlighted as important (with a positive sign).
    \item The density of the chambers and aorta are highlighted as important (with a negative sign).
    \item The density features of the fatty tissues are highlighted as important (with a positive sign).
\end{itemize}

The main findings from the saliency analysis for age prediction are:
\begin{itemize}
    \item The volume of the RA, and aorta (both the descending aorta and the ascending aorta which is not visible in the presented slice but can be seen in the full saliency map volume) show positive associations with age.
    \item The volume of the myocardium shows an overall negative association with age.
    \item The density of the myocardium shows an overall positive association with age.
    \item The density of the fatty tissue has a negative association with age.
    \item Females have a positive relationship between volume and age in the apex, while males do not.
    \item Females have a strong positive association between the volume of the LA and age while there is no correlation between the volume of the LA and age in males.
    \item Both the density and volume of the tissue surrounding the descending aorta have a positive association with age.
    \item The Stddev maps highlight the areas containing and surrounding the valves, and the boundary of the LV, as well as a low to medium overall saliency in the chambers.
\end{itemize}

%\clearpage

\section{Discussion}

The proof-of-concept experiments demonstrated that the spatially standardized supervoxel-based features do contain sufficiently detailed information about the morphology of the heart to accurately predict the volumes of key morphological sub-structures. Furthermore, the relevance of the saliency maps is confirmed by observing that known regions of relevance are highlighted as important, and with a positive association as expected.

For reasons of simplicity and interpretability, it is common to use measurements derived from segmentation as features for aging studies and other regression tasks \cite{pickhardt2024novel,wasserthal2023totalsegmentator}, where (typically coarse-grained) known regions of importance are segmented and features are extracted from them, such as mean intensity or volume. On the other hand, the supervoxel-based representations are not derived from prior knowledge about relevant regions, but are rather derived from the image contents, and are substantially more fine-grained than what off-the-shelf methods such as TotalSegmentator provides \cite{wasserthal2023totalsegmentator}. Through the use of linear models and saliency analysis, the interpretability advantage of the segmentation-based methods can be preserved, while a substantially higher prediction performance is possible, which was observed in a comparative study between the supervoxel-based features and segmentation-derived features (incorporating 12 TotalSegmentator-derived features of mean density and volume of the chambers, myocardium, and aorta).

%We observed that the supervoxel-based features, which provide a detailed representation of the local morphology in all sub-regions of the heart provided substantially more accurate age prediction performance compared to that achieved with selected segmented coarse-grained sub-regions representing the chambers, the myocardium and the aorta. Through the use of linear models and saliency analysis, interpretability is preserved, while substantially higher performance in the predictions can be achieved.
The main associations found in this work agree with the supervoxel-wise correlation and segmentation-based correlation in \cite{ofverstedt2024method}, which was performed on the same dataset. The segmentation-based findings in \cite{wasserthal2023totalsegmentator} also agree well with the signs of the associations between the volumes of the chambers and aorta and age. The main additional knowledge is the extent to which the age can be predicted from the \emph{whole heart} or from sub-regions, combining multiple features from numerous supervoxels, rather than investigating the correlation between single features and age.

The performance reported in different age prediction studies tends to not be directly comparable, and is dependent on factors such as imaging aspects (the image modality, image resolution, level of detail, and field-of-view), as well as characteristics of the cohort examined, such as the number of subjects, the range (and distribution) of ages, and also health status (the presence of disease or anatomical abnormalities). The more informative aspects of this study are therefore relative performance between different ROIs, and the agreement between those predictions, rather than the absolute levels of performance. The performance in terms of $\mathtt{MAE}$ is comparable with some earlier studies \cite{langner2019identifying,starck2023atlas}, but those studies were conducted on MRI datasets and with wider age ranges. On the other hand, one study conducted on chest CT images \cite{kerber2023deep} did observe much lower performance than in this study, with a $\mathtt{MAE}=5.76$, though the study was performed on a dataset with more variability in the imaging protocol, and health status of the subjects, as well as a wider age range.

There is strong agreement between the predicted ages from different sub-regions and also between predictions based on different subsets of the four supervoxel features. We observed a higher agreement between the predictions from the \emph{whole heart} and each other region than between the predicted ages and the chronological age, despite having used chronological age to fit the models. This is likely related to the correlation of the features (between density and volume, which can be seen as associations between density features and volume in the proof-of-concept study saliency maps), as well as between different sub-regions (which can be seen as a weak positive association to LVV everywhere in the heart). In particular, the models with the highest performance also agreed to the highest extent. The inter-region and inter-feature agreement provides evidence for both the plausibility of the results and the stability of the predicted ages. High agreement in age prediction between different ROIs has been observed previously for different organs (including the heart) in MRI images \cite{ecker2024deep}.

\subsection{Limitations}

Although linear models exhibited comparable (or better) performance than more complex non-linear models, this is likely an effect of the size of the dataset and the limited age range. Linear models are limited in their power to capture the relationship between the features and the target variable, and we hypothesize that a linear model may not perform better than e.g. deep models for substantially wider age ranges, where non-linear relationships may be crucial. Linear models lack the ability to find more complex patterns across features such as curvature in the feature/target relationship, multiplicative effects, and high-level feature interactions. 

The inter-subject registration of different individuals into a common space is not an error-free process \cite{ofverstedt2024method}, and factors such as anatomical differences, artifacts, differences in the value distribution of the CT images, and contrast differences can all impact the results. Registration errors introduce noise in the dataset due to the partial overlap of different tissue types in a single supervoxel, and due to noise in the local volume measures. These errors weaken the relationship between the features and the targets in addition to any inherent variability of morphology for a given chronological age.

Regression dilution/attenuation, which is a well-documented phenomenon where the presence of noise and measurement errors in the independent variables causes an underestimation of the regression slopes \cite{frost2000correcting}, making it challenging to define a meaningful age gap \cite{ecker2024deep}, i.e. the difference between predicted ages and chronological ages, or to infer a biological age that relates to a subject's health status compared to a typical healthy individual of that particular chronological age. However, for the goal of using the predicted ages as predictors of disease and clinical outcomes jointly with the chronological age, the regression dilution is likely not a limiting factor.

The highest $R^2$ scores achieved in the chronological age evaluation are not very high, which is likely caused by a combination of (i) the limited age span (50-65 years), (ii) the amount of information present in the images to relate the morphology to the chronological age, and (iii) the dilution factor of the biological age that relates to the subject’s health.

Despite the level of detail captured by the spatially standardized supervoxel-based features, some aspects of the original images may still not be fully represented, such as the presence or absence of plaque in the coronary vessels (vessels which were the most challenging structures to register between subjects \cite{ofverstedt2024method}) and narrowing of the vessels. For a more complete analysis, future work could aim to detect such features and incorporate their presence or absence into the regression, alongside the supervoxel-based features.

\section{Conclusion}

We developed a feature extraction and machine learning method based on spatially standardized supervoxels for age prediction from CTA images. The method was applied to a subset of SCAPIS with observed errors (using the chronological age as reference) of $\mathtt{MAE}=2.74$ and $R^2=0.436$ for females ($n=721$), and $\mathtt{MAE}=2.78$ and $R^2=0.381$ for males ($n=666$). 

We also studied the pairwise correlations between the predicted ages for various subregions of the heart and between each prediction ROI and the chronological age. In comparison to the prediction of chronological age, we observed higher agreement between the age predictions of the whole heart and different other sub-regions of the heart. This observed stability of these predicted ages suggests that we may view them as a \emph{morphological heart age} biomarker. 

Future work includes application to the full SCAPIS dataset when that is made available, as well as investigating the clinical relevance of this morphological heart age in terms of association to common risk factors for disease, and as a predictor for disease and survival.

\section*{Acknowledgements}

This research was funded by the Swedish Heart and Lung Foundation through grant number 2022012924, and EXODIAB. 

The main funding body of The Swedish CArdioPulmonary bioImage Study (SCAPIS) is the Swedish Heart-Lung Foundation. The study is also funded by the Knut and Alice Wallenberg Foundation, the Swedish Research Council and VINNOVA (Sweden’s Innovation agency) the University of Gothenburg and Sahlgrenska University Hospital, Karolinska Institutet and Stockholm county council, Link\"{o}ping University and University Hospital, Lund University and Sk{\aa}ne University Hospital, Ume{\aa} University and University Hospital, Uppsala University and University Hospital.

%\appendix
%\section{My Appendix}
%Appendix sections are coded under \verb+\appendix+.

%\verb+\printcredits+ command is used after appendix sections to list 
%author credit taxonomy contribution roles tagged using \verb+\credit+ 
%in frontmatter.
\printcredits

%% Loading bibliography style file
\bibliographystyle{model1-num-names}
%\bibliographystyle{cas-model2-names}

% Loading bibliography database
\bibliography{references}

\begin{thebibliography}{22}
\expandafter\ifx\csname natexlab\endcsname\relax\def\natexlab#1{#1}\fi
\providecommand{\url}[1]{\texttt{#1}}
\providecommand{\href}[2]{#2}
\providecommand{\path}[1]{#1}
\providecommand{\DOIprefix}{doi:}
\providecommand{\ArXivprefix}{arXiv:}
\providecommand{\URLprefix}{URL: }
\providecommand{\Pubmedprefix}{pmid:}
\providecommand{\doi}[1]{\href{http://dx.doi.org/#1}{\path{#1}}}
\providecommand{\Pubmed}[1]{\href{pmid:#1}{\path{#1}}}
\providecommand{\bibinfo}[2]{#2}
\ifx\xfnm\relax \def\xfnm[#1]{\unskip,\space#1}\fi
%Type = Article
\bibitem[{Langner et~al.(2019)Langner, Wikstr{\"o}m, Bjerner, Ahlstr{\"o}m, and Kullberg}]{langner2019identifying}
\bibinfo{author}{T.~Langner}, \bibinfo{author}{J.~Wikstr{\"o}m}, \bibinfo{author}{T.~Bjerner}, \bibinfo{author}{H.~Ahlstr{\"o}m}, \bibinfo{author}{J.~Kullberg},
\newblock \bibinfo{title}{Identifying morphological indicators of aging with neural networks on large-scale whole-body mri},
\newblock \bibinfo{journal}{IEEE transactions on medical imaging} \bibinfo{volume}{39} (\bibinfo{year}{2019}) \bibinfo{pages}{1430--1437}.
%Type = Inproceedings
\bibitem[{Ecker et~al.(2024)Ecker, Fr{\"u}h, Yang, Gatidis, and K{\"u}stner}]{ecker2024deep}
\bibinfo{author}{V.~Ecker}, \bibinfo{author}{M.~Fr{\"u}h}, \bibinfo{author}{B.~Yang}, \bibinfo{author}{S.~Gatidis}, \bibinfo{author}{T.~K{\"u}stner},
\newblock \bibinfo{title}{Deep regression for biological age estimation in multiple organs: Investigations on 40,000 subjects of the uk biobank},
\newblock in: \bibinfo{booktitle}{ICASSP 2024-2024 IEEE International Conference on Acoustics, Speech and Signal Processing (ICASSP)}, \bibinfo{organization}{IEEE}, \bibinfo{year}{2024}, pp. \bibinfo{pages}{2255--2259}.
%Type = Article
\bibitem[{Sihag et~al.(2024)Sihag, Mateos, McMillan, and Ribeiro}]{sihag2024explainable}
\bibinfo{author}{S.~Sihag}, \bibinfo{author}{G.~Mateos}, \bibinfo{author}{C.~McMillan}, \bibinfo{author}{A.~Ribeiro},
\newblock \bibinfo{title}{Explainable brain age prediction using covariance neural networks},
\newblock \bibinfo{journal}{Advances in Neural Information Processing Systems} \bibinfo{volume}{36} (\bibinfo{year}{2024}).
%Type = Misc
\bibitem[{Starck et~al.(2023)Starck, Kini, Maria, Ritter, R{\"u}ckert, and M{\"u}ller}]{starck2023atlas}
\bibinfo{author}{S.~Starck}, \bibinfo{author}{Y.~V. Kini}, \bibinfo{author}{J.~J. Maria}, \bibinfo{author}{R.~B. Ritter}, \bibinfo{author}{D.~R{\"u}ckert}, \bibinfo{author}{T.~M{\"u}ller}, \bibinfo{title}{Atlas-based interpretable age prediction in whole-body mr images}, \bibinfo{year}{2023}.
%Type = Article
\bibitem[{Kerber et~al.(2023)Kerber, Hepp, K{\"u}stner, and Gatidis}]{kerber2023deep}
\bibinfo{author}{B.~Kerber}, \bibinfo{author}{T.~Hepp}, \bibinfo{author}{T.~K{\"u}stner}, \bibinfo{author}{S.~Gatidis},
\newblock \bibinfo{title}{Deep learning-based age estimation from clinical computed tomography image data of the thorax and abdomen in the adult population},
\newblock \bibinfo{journal}{Plos one} \bibinfo{volume}{18} (\bibinfo{year}{2023}) \bibinfo{pages}{e0292993}.
%Type = Article
\bibitem[{Pickhardt et~al.(2024)Pickhardt, Kattan, Lee, Pooler, Pyrros, Liu, Zea, Summers, and Garrett}]{pickhardt2024novel}
\bibinfo{author}{P.~Pickhardt}, \bibinfo{author}{M.~Kattan}, \bibinfo{author}{M.~Lee}, \bibinfo{author}{B.~D. Pooler}, \bibinfo{author}{A.~Pyrros}, \bibinfo{author}{D.~Liu}, \bibinfo{author}{R.~Zea}, \bibinfo{author}{R.~Summers}, \bibinfo{author}{J.~Garrett},
\newblock \bibinfo{title}{Novel biological age model using explainable automated ct-based cardiometabolic biomarkers for phenotypic prediction of longevity}  (\bibinfo{year}{2024}).
%Type = Article
\bibitem[{Lima et~al.(2021)Lima, Ribeiro, Paix{\~a}o, Ribeiro, Pinto-Filho, Gomes, Oliveira, Sabino, Duncan, Giatti et~al.}]{lima2021deep}
\bibinfo{author}{E.~M. Lima}, \bibinfo{author}{A.~H. Ribeiro}, \bibinfo{author}{G.~M. Paix{\~a}o}, \bibinfo{author}{M.~H. Ribeiro}, \bibinfo{author}{M.~M. Pinto-Filho}, \bibinfo{author}{P.~R. Gomes}, \bibinfo{author}{D.~M. Oliveira}, \bibinfo{author}{E.~C. Sabino}, \bibinfo{author}{B.~B. Duncan}, \bibinfo{author}{L.~Giatti}, et~al.,
\newblock \bibinfo{title}{Deep neural network-estimated electrocardiographic age as a mortality predictor},
\newblock \bibinfo{journal}{Nature communications} \bibinfo{volume}{12} (\bibinfo{year}{2021}) \bibinfo{pages}{5117}.
%Type = Article
\bibitem[{Min et~al.(2010)Min, Shaw, and Berman}]{min2010present}
\bibinfo{author}{J.~K. Min}, \bibinfo{author}{L.~J. Shaw}, \bibinfo{author}{D.~S. Berman},
\newblock \bibinfo{title}{The present state of coronary computed tomography angiography: a process in evolution},
\newblock \bibinfo{journal}{Journal of the American College of Cardiology} \bibinfo{volume}{55} (\bibinfo{year}{2010}) \bibinfo{pages}{957--965}.
%Type = Article
\bibitem[{Bergstr{\"o}m et~al.(2015)Bergstr{\"o}m, Berglund, Blomberg, Brandberg, Engstr{\"o}m, Engvall, Eriksson, De~Faire, Flinck, Hansson et~al.}]{bergstrom2015swedish}
\bibinfo{author}{G.~Bergstr{\"o}m}, \bibinfo{author}{G.~Berglund}, \bibinfo{author}{A.~Blomberg}, \bibinfo{author}{J.~Brandberg}, \bibinfo{author}{G.~Engstr{\"o}m}, \bibinfo{author}{J.~Engvall}, \bibinfo{author}{M.~Eriksson}, \bibinfo{author}{U.~De~Faire}, \bibinfo{author}{A.~Flinck}, \bibinfo{author}{M.~G. Hansson}, et~al.,
\newblock \bibinfo{title}{The swedish cardiopulmonary bioimage study: objectives and design},
\newblock \bibinfo{journal}{Journal of internal medicine} \bibinfo{volume}{278} (\bibinfo{year}{2015}) \bibinfo{pages}{645--659}.
%Type = Article
\bibitem[{Chen et~al.(2020)Chen, Qin, Qiu, Tarroni, Duan, Bai, and Rueckert}]{chen2020deep}
\bibinfo{author}{C.~Chen}, \bibinfo{author}{C.~Qin}, \bibinfo{author}{H.~Qiu}, \bibinfo{author}{G.~Tarroni}, \bibinfo{author}{J.~Duan}, \bibinfo{author}{W.~Bai}, \bibinfo{author}{D.~Rueckert},
\newblock \bibinfo{title}{Deep learning for cardiac image segmentation: a review},
\newblock \bibinfo{journal}{Frontiers in Cardiovascular Medicine} \bibinfo{volume}{7} (\bibinfo{year}{2020}) \bibinfo{pages}{25}.
%Type = Article
\bibitem[{Wasserthal et~al.(2023)Wasserthal, Breit, Meyer, Pradella, Hinck, Sauter, Heye, Boll, Cyriac, Yang et~al.}]{wasserthal2023totalsegmentator}
\bibinfo{author}{J.~Wasserthal}, \bibinfo{author}{H.-C. Breit}, \bibinfo{author}{M.~T. Meyer}, \bibinfo{author}{M.~Pradella}, \bibinfo{author}{D.~Hinck}, \bibinfo{author}{A.~W. Sauter}, \bibinfo{author}{T.~Heye}, \bibinfo{author}{D.~T. Boll}, \bibinfo{author}{J.~Cyriac}, \bibinfo{author}{S.~Yang}, et~al.,
\newblock \bibinfo{title}{Totalsegmentator: Robust segmentation of 104 anatomic structures in ct images},
\newblock \bibinfo{journal}{Radiology: Artificial Intelligence} \bibinfo{volume}{5} (\bibinfo{year}{2023}).
%Type = Article
\bibitem[{{\"O}fverstedt et~al.(2024){\"O}fverstedt, Lundstr{\"o}m, Bergstr{\"o}m, Kullberg, and Ahlstr{\"o}m}]{ofverstedt2024method}
\bibinfo{author}{J.~{\"O}fverstedt}, \bibinfo{author}{E.~Lundstr{\"o}m}, \bibinfo{author}{G.~Bergstr{\"o}m}, \bibinfo{author}{J.~Kullberg}, \bibinfo{author}{H.~Ahlstr{\"o}m},
\newblock \bibinfo{title}{A method for supervoxel-wise association studies of age and other non-imaging variables from coronary computed tomography angiograms},
\newblock \bibinfo{journal}{arXiv preprint arXiv:2405.07762}  (\bibinfo{year}{2024}).
%Type = Inproceedings
\bibitem[{Tang and Chung(2007)}]{tang2007non}
\bibinfo{author}{T.~W. Tang}, \bibinfo{author}{A.~C. Chung},
\newblock \bibinfo{title}{Non-rigid image registration using graph-cuts},
\newblock in: \bibinfo{booktitle}{Medical Image Computing and Computer-Assisted Intervention--MICCAI 2007: 10th International Conference, Brisbane, Australia, October 29-November 2, 2007, Proceedings, Part I 10}, \bibinfo{organization}{Springer}, \bibinfo{year}{2007}, pp. \bibinfo{pages}{916--924}.
%Type = Article
\bibitem[{Ekstr{\"o}m et~al.(2020)Ekstr{\"o}m, Malmberg, Ahlstr{\"o}m, Kullberg, and Strand}]{ekstrom2020fast}
\bibinfo{author}{S.~Ekstr{\"o}m}, \bibinfo{author}{F.~Malmberg}, \bibinfo{author}{H.~Ahlstr{\"o}m}, \bibinfo{author}{J.~Kullberg}, \bibinfo{author}{R.~Strand},
\newblock \bibinfo{title}{Fast graph-cut based optimization for practical dense deformable registration of volume images},
\newblock \bibinfo{journal}{Computerized Medical Imaging and Graphics} \bibinfo{volume}{84} (\bibinfo{year}{2020}) \bibinfo{pages}{101745}.
%Type = Article
\bibitem[{Ekstr{\"o}m et~al.(2021)Ekstr{\"o}m, Pilia, Kullberg, Ahlstr{\"o}m, Strand, and Malmberg}]{ekstrom2021faster}
\bibinfo{author}{S.~Ekstr{\"o}m}, \bibinfo{author}{M.~Pilia}, \bibinfo{author}{J.~Kullberg}, \bibinfo{author}{H.~Ahlstr{\"o}m}, \bibinfo{author}{R.~Strand}, \bibinfo{author}{F.~Malmberg},
\newblock \bibinfo{title}{Faster dense deformable image registration by utilizing both cpu and gpu},
\newblock \bibinfo{journal}{Journal of Medical Imaging} \bibinfo{volume}{8} (\bibinfo{year}{2021}) \bibinfo{pages}{014002--014002}.
%Type = Article
\bibitem[{J{\"o}nsson et~al.(2022)J{\"o}nsson, Ekstr{\"o}m, Strand, Pedersen, Molin, Ahlstr{\"o}m, and Kullberg}]{jonsson2022image}
\bibinfo{author}{H.~J{\"o}nsson}, \bibinfo{author}{S.~Ekstr{\"o}m}, \bibinfo{author}{R.~Strand}, \bibinfo{author}{M.~A. Pedersen}, \bibinfo{author}{D.~Molin}, \bibinfo{author}{H.~Ahlstr{\"o}m}, \bibinfo{author}{J.~Kullberg},
\newblock \bibinfo{title}{An image registration method for voxel-wise analysis of whole-body oncological pet-ct},
\newblock \bibinfo{journal}{Scientific Reports} \bibinfo{volume}{12} (\bibinfo{year}{2022}) \bibinfo{pages}{18768}.
%Type = Article
\bibitem[{Achanta et~al.(2012)Achanta, Shaji, Smith, Lucchi, Fua, and S{\"u}sstrunk}]{achanta2012slic}
\bibinfo{author}{R.~Achanta}, \bibinfo{author}{A.~Shaji}, \bibinfo{author}{K.~Smith}, \bibinfo{author}{A.~Lucchi}, \bibinfo{author}{P.~Fua}, \bibinfo{author}{S.~S{\"u}sstrunk},
\newblock \bibinfo{title}{Slic superpixels compared to state-of-the-art superpixel methods},
\newblock \bibinfo{journal}{IEEE transactions on pattern analysis and machine intelligence} \bibinfo{volume}{34} (\bibinfo{year}{2012}) \bibinfo{pages}{2274--2282}.
%Type = Article
\bibitem[{Beare et~al.(2018)Beare, Lowekamp, and Yaniv}]{beare2018image}
\bibinfo{author}{R.~Beare}, \bibinfo{author}{B.~Lowekamp}, \bibinfo{author}{Z.~Yaniv},
\newblock \bibinfo{title}{Image segmentation, registration and characterization in r with simpleitk},
\newblock \bibinfo{journal}{Journal of statistical software} \bibinfo{volume}{86} (\bibinfo{year}{2018}).
%Type = Article
\bibitem[{Pedregosa et~al.(2011)Pedregosa, Varoquaux, Gramfort, Michel, Thirion, Grisel, Blondel, Prettenhofer, Weiss, Dubourg, Vanderplas, Passos, Cournapeau, Brucher, Perrot, and Duchesnay}]{scikit-learn}
\bibinfo{author}{F.~Pedregosa}, \bibinfo{author}{G.~Varoquaux}, \bibinfo{author}{A.~Gramfort}, \bibinfo{author}{V.~Michel}, \bibinfo{author}{B.~Thirion}, \bibinfo{author}{O.~Grisel}, \bibinfo{author}{M.~Blondel}, \bibinfo{author}{P.~Prettenhofer}, \bibinfo{author}{R.~Weiss}, \bibinfo{author}{V.~Dubourg}, \bibinfo{author}{J.~Vanderplas}, \bibinfo{author}{A.~Passos}, \bibinfo{author}{D.~Cournapeau}, \bibinfo{author}{M.~Brucher}, \bibinfo{author}{M.~Perrot}, \bibinfo{author}{E.~Duchesnay},
\newblock \bibinfo{title}{Scikit-learn: Machine learning in {P}ython},
\newblock \bibinfo{journal}{Journal of Machine Learning Research} \bibinfo{volume}{12} (\bibinfo{year}{2011}) \bibinfo{pages}{2825--2830}.
%Type = Article
\bibitem[{Sirovich and Kirby(1987)}]{sirovich1987low}
\bibinfo{author}{L.~Sirovich}, \bibinfo{author}{M.~Kirby},
\newblock \bibinfo{title}{Low-dimensional procedure for the characterization of human faces},
\newblock \bibinfo{journal}{Josa a} \bibinfo{volume}{4} (\bibinfo{year}{1987}) \bibinfo{pages}{519--524}.
%Type = Misc
\bibitem[{Öfverstedt(2024)}]{ofverstedt_2024_13832059}
\bibinfo{author}{J.~Öfverstedt}, \bibinfo{title}{{Morphological heart age from CTA: Predictions, Performance and Saliency Maps}}, \bibinfo{year}{2024}. \URLprefix \url{https://doi.org/10.5281/zenodo.13832059}. \DOIprefix\doi{10.5281/zenodo.13832059}.
%Type = Article
\bibitem[{Frost and Thompson(2000)}]{frost2000correcting}
\bibinfo{author}{C.~Frost}, \bibinfo{author}{S.~G. Thompson},
\newblock \bibinfo{title}{Correcting for regression dilution bias: comparison of methods for a single predictor variable},
\newblock \bibinfo{journal}{Journal of the Royal Statistical Society Series A: Statistics in Society} \bibinfo{volume}{163} (\bibinfo{year}{2000}) \bibinfo{pages}{173--189}.

\end{thebibliography}

\appendix
\section{Feature correlation}

\begin{figure}[ht]
    \centering
    \includegraphics[width=0.95\linewidth]{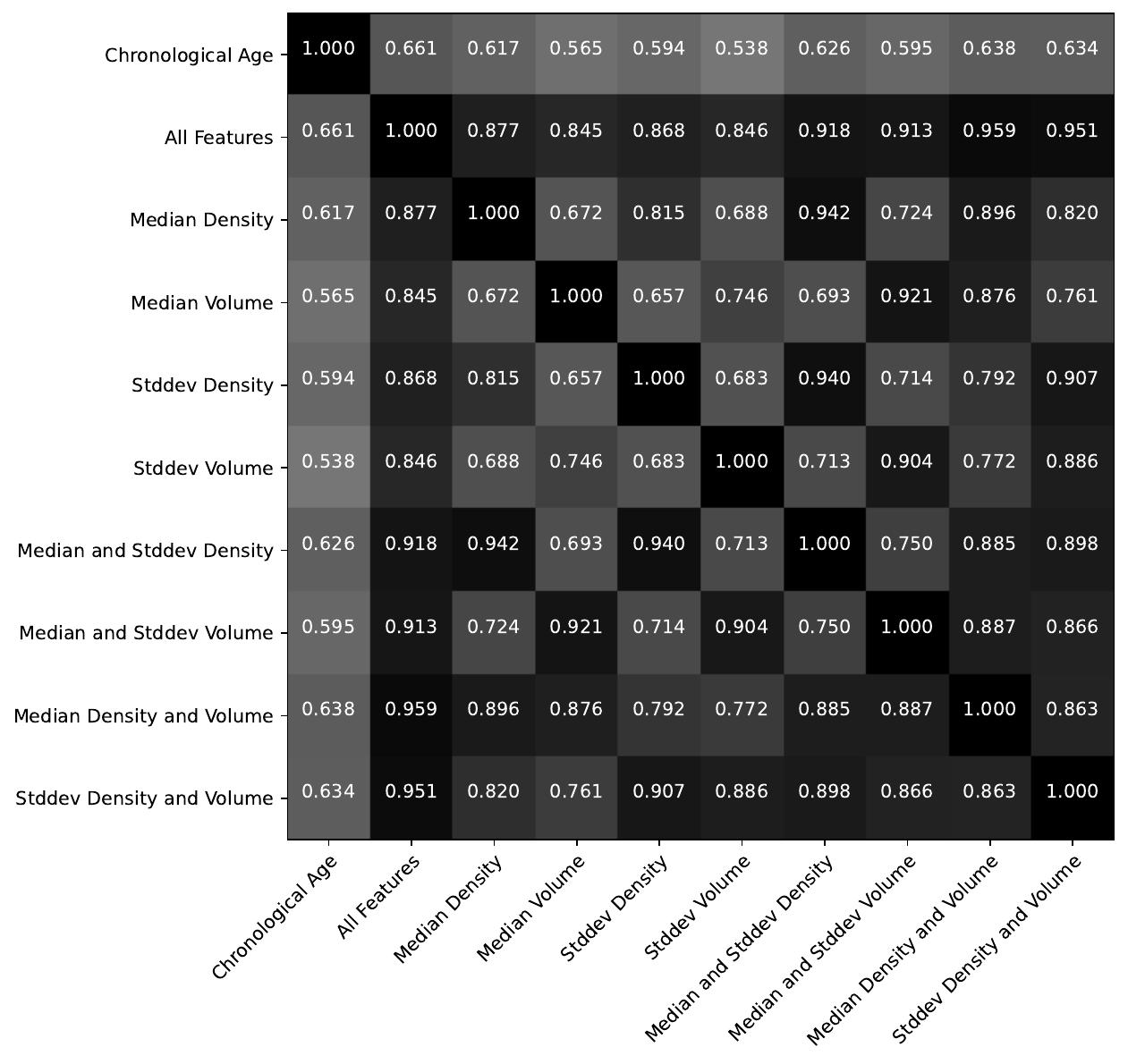}
    \caption{Pearson correlation between the age prediction of various supervoxel feature combinations, as well as chronological age, for the female sub-group.}
    \label{fig:featurecorrelationfemale}
\end{figure}

\begin{figure}[ht]
    \centering
    \includegraphics[width=0.95\linewidth]{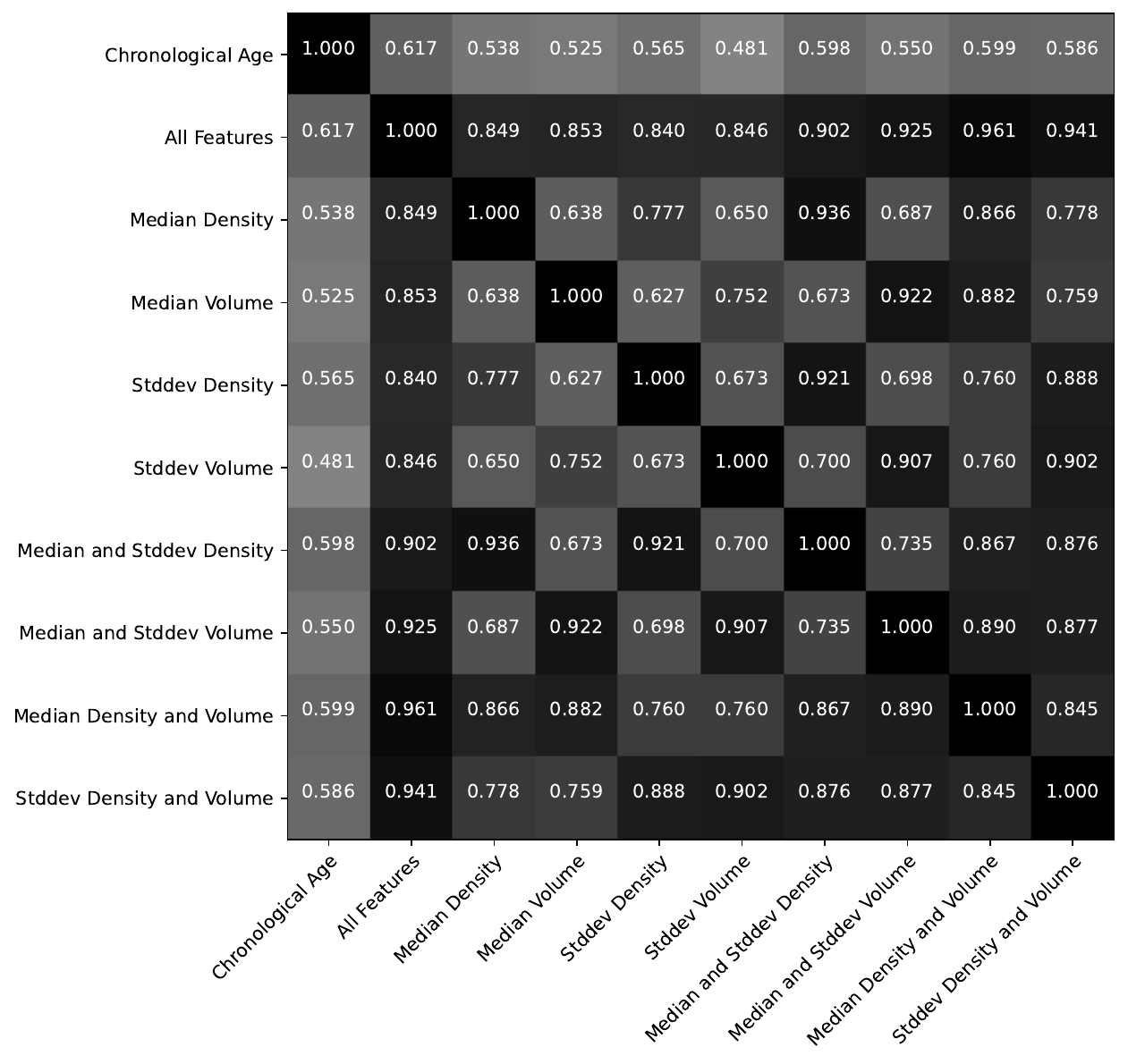}
    \caption{Pearson correlation between the age prediction of various supervoxel feature combinations, as well as chronological age, for the male sub-group.}
    \label{fig:featurecorrelationmale}
\end{figure}

The correlation between models using \emph{all features} and the other feature combinations are explored in figures Fig.~\ref{fig:featurecorrelationfemale} for females, and Fig.~\ref{fig:featurecorrelationmale} for males. We observed a high correlation between the predictions using \emph{all features} and each other feature combination.

We also present the correlation between the chronological age, an the predicted age from the \emph{whole heart}, with the volumetric features LVV, RVV, LAV, RAV, MYOV, AV, in Tab.~\ref{tab:agevolcorrelation}.

\begin{table}[]
    \centering
    \caption{Pearson correlation between age (chronological and predicted), and 6 explicit volume measurements. The correlations between age and the volumetric features strengthen for features with a high correlation to chronological age.}
    \label{tab:agevolcorrelation}
    %\resizebox{\linewidth}{!}{
    \begin{tabular}{c|c|c|c|c|c|c|c}
        Age & Sex & LVV & RVV & LAV & RAV & MYOV & AV \\ \hline \hline
        \multirow{2}{*}{Chronological age} & Female & $-0.039$ & $-0.068$ & $0.259$ & $0.103$ & $0.010$ & $0.229$ \\
        & Male & $-0.181$ & $-0.192$ & $0.018$ & $-0.025$ & $-0.133$ & $0.160$ \\ \hline \hline
        Predicted age from & Female & $-0.030$ & $-0.076$ & $0.364$ & $0.128$ & $0.042$ & $0.337$ \\
        the whole heart & Male & $-0.275$ & $-0.302$ & $0.018$ & $-0.057$ & $-0.201$ & $0.192$ \\ \hline \hline
    \end{tabular}%}
\end{table}

\section{Saliency analysis}

\begin{figure}
    \centering
    \textbf{Saliency maps for age prediction without PCA dimensionality reduction (FEMALE)}\\
    \subfloat[][Median density]{
    \includegraphics[width=0.24\linewidth]{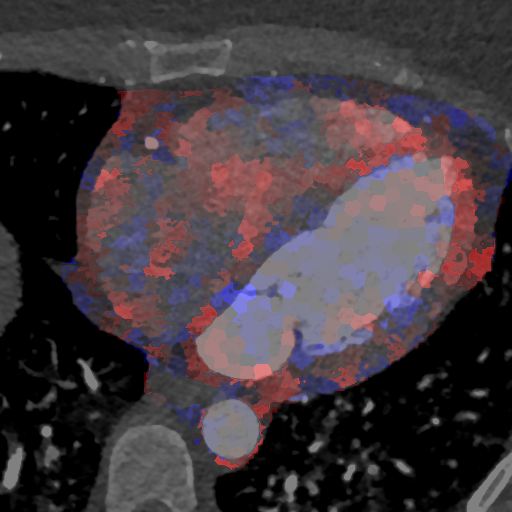}}
    \subfloat[][Median volume]{
    \includegraphics[width=0.24\linewidth]{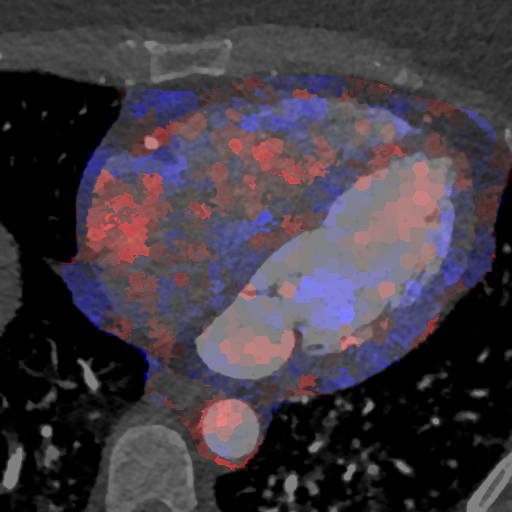}}
    \subfloat[][Stddev density]{
    \includegraphics[width=0.24\linewidth]{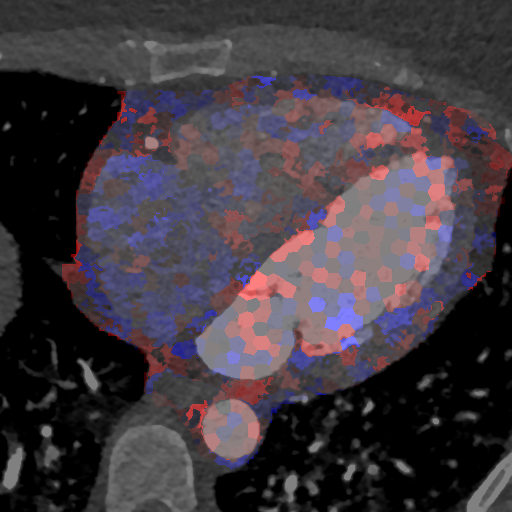}}
    \subfloat[][Stddev volume]{
    \includegraphics[width=0.24\linewidth]{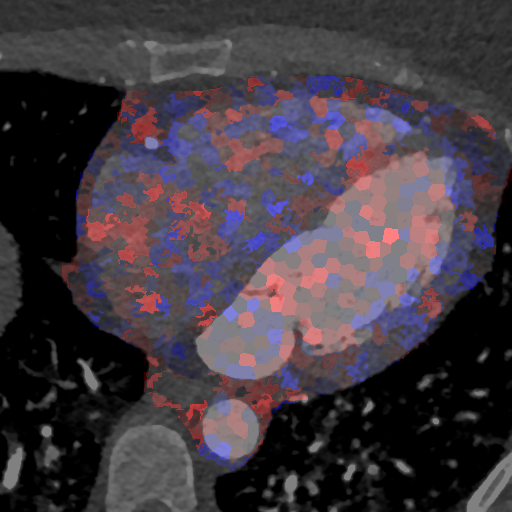}}\\
    \vspace{8mm}
    \textbf{Saliency maps for age prediction with PCA dimensionality reduction (FEMALE)}\\
    \subfloat[][Median density]{
    \includegraphics[width=0.24\linewidth]{figs/newsaliency/female_median_density.png}}
    \subfloat[][Median volume]{
    \includegraphics[width=0.24\linewidth]{figs/newsaliency/female_median_volume.png}}
    \subfloat[][Stddev density]{
    \includegraphics[width=0.24\linewidth]{figs/newsaliency_lvv/lvv_female_stddev_density.png}}
    \subfloat[][Stddev volume]{
    \includegraphics[width=0.24\linewidth]{figs/newsaliency/female_stddev_volume.png}}\\
    \hfill\includegraphics[width=0.3\linewidth]{figs/saliency_colormap.pdf}    
    \caption{An axial slice of the saliency map for the age prediction for females without PCA dimensionality reduction (a-d) and with PCA dimensionality reduction (e-h). Compared to the saliency maps with PCA the associations detected are much more noisy.}
    \label{fig:orig_saliency_female_densvol}
\end{figure}

Figure \ref{fig:orig_saliency_female_densvol} displays saliency maps for age prediction excluding the PCA dimensionality reduction to investigate the impact of that computational step on the saliency analysis. We observed overall similar results, but many more noisy associations (red spots within predominately blue areas, and blue spots within predominately red areas).

\section{Appendix C: Ablation experiments}

\begin{figure}
    \centering
    \subfloat[][Grid size performance study (Female)]{
    \includegraphics[width=0.49\linewidth]{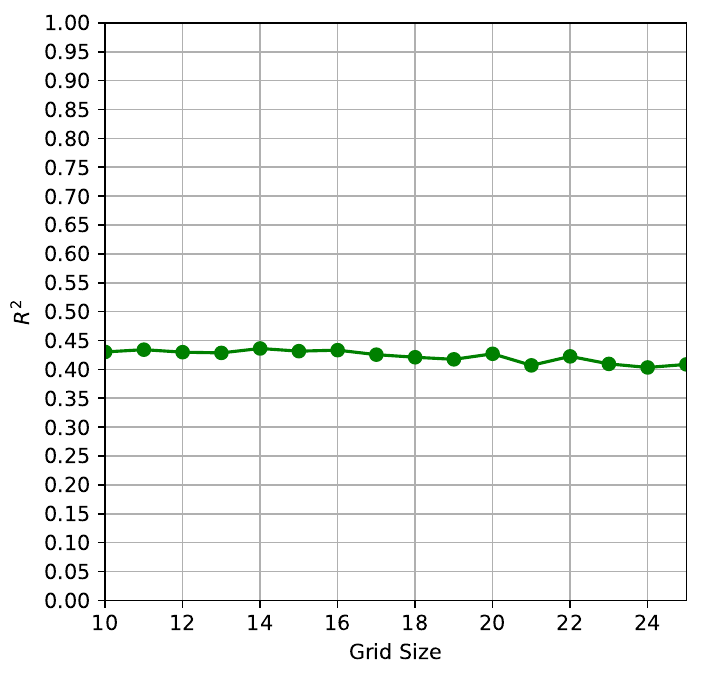}}
    \subfloat[][Grid size performance study (Male)]{
    \includegraphics[width=0.49\linewidth]{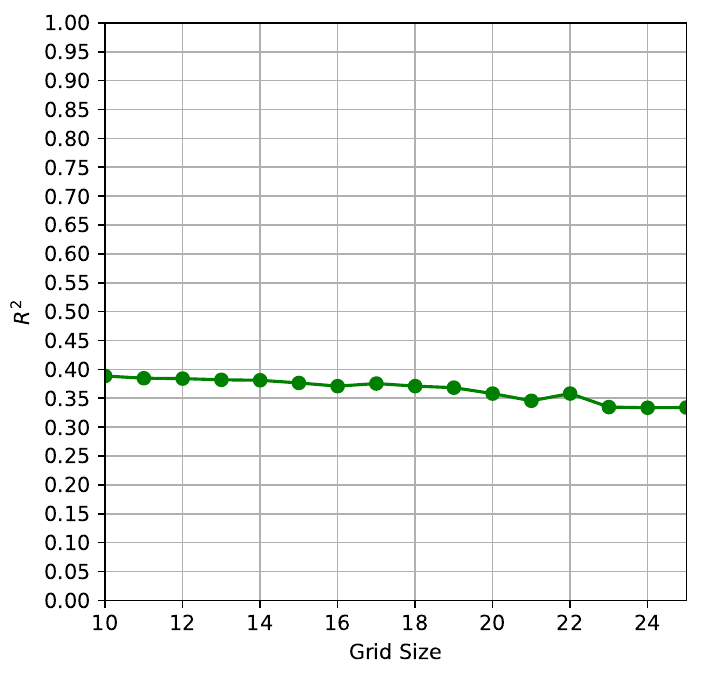}}
    \caption{Performance (in $R^2$) of the age prediction as a function of the \emph{grid size} parameter.}
    \label{fig:ablation_graph_grid_size}
\end{figure}

\begin{figure}
    \centering
    \subfloat[][Clip level performance study (Female)]{
    \includegraphics[width=0.49\linewidth]{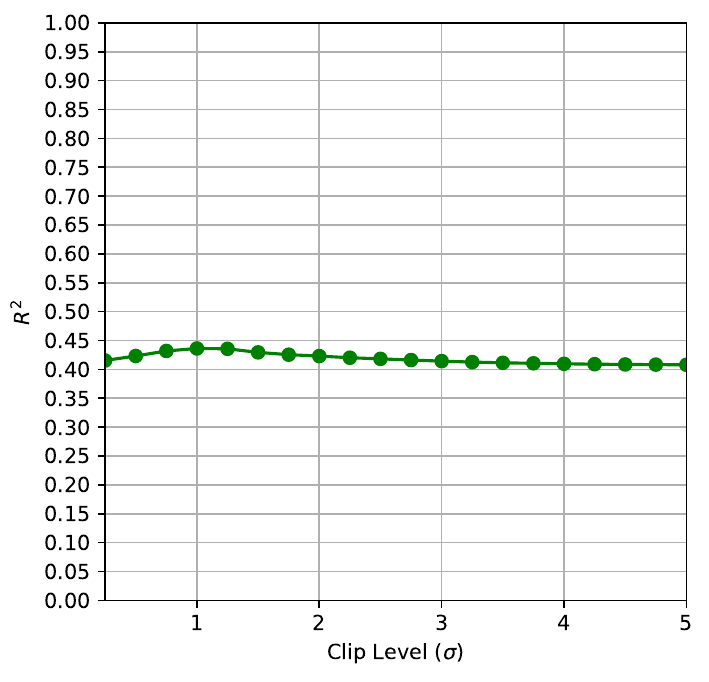}}
    \subfloat[][Clip level performance study (Male)]{
    \includegraphics[width=0.49\linewidth]{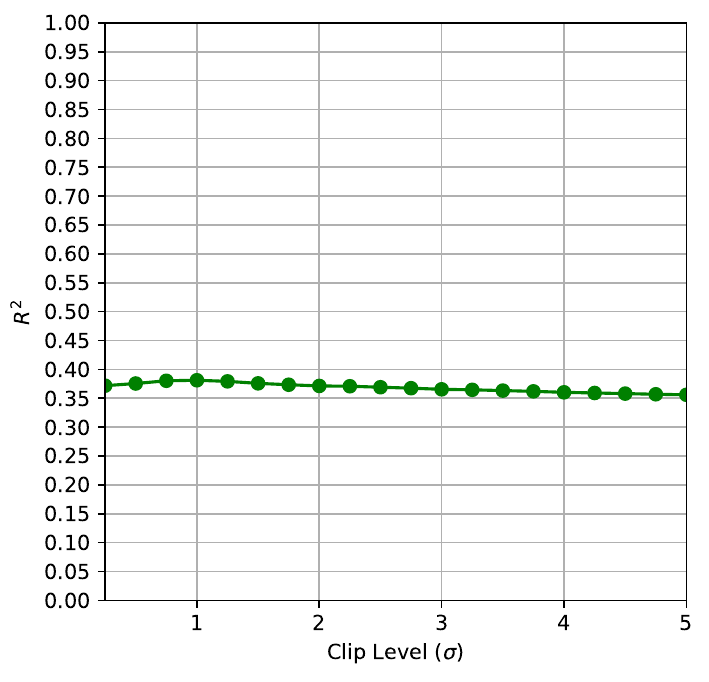}}
    \caption{Performance (in $R^2$) of the age prediction as a function of the \emph{clip level} parameter.}
    \label{fig:ablation_graph_clip_level}
\end{figure}

Here we present the results of the ablation experiments for (i) the supervoxel size controlled by the \emph{grid size} parameter, shown in Fig~\ref{fig:ablation_graph_grid_size}, and (ii) the clip level in standard deviations used to truncate the input features before the PCA is applied, shown in Fig~\ref{fig:ablation_graph_clip_level}.

The best observed \emph{clip level} is 1.0 in both females and males. To determine the best grid size, the relation between the parameter and the performance is more complicated, with differences between the sexes. For males, the best performance (in terms of $R^2$) is achieved at grid size 10, while the best value for females is found at $14$. An overall good performance can be achieved at a grid size of $14$, and that is what is used in the remaining experiments.

\begin{figure}
    \centering
    \subfloat[][PCA components study (Female)]{
    \includegraphics[width=0.49\linewidth]{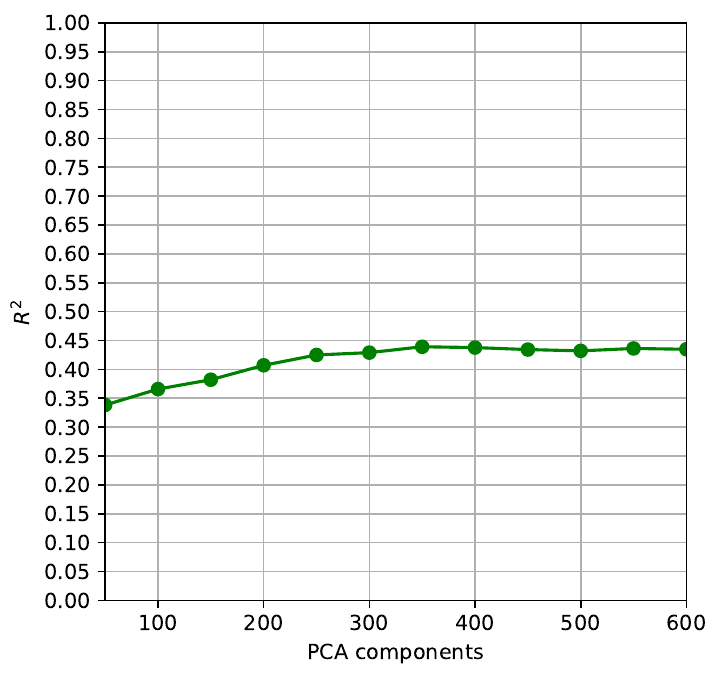}}
    \subfloat[][PCA components study (Male)]{
    \includegraphics[width=0.49\linewidth]{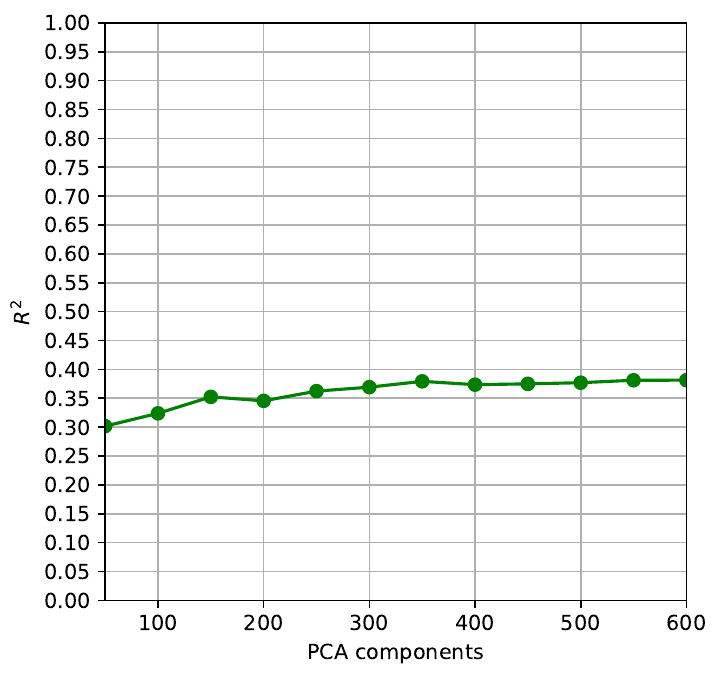}}
    \caption{Performance (in $R^2$) of the age prediction as a function of the number of PCA components for the \emph{whole heart} ROI.}
    \label{fig:ablation_graph_pca_n}
\end{figure}

\section{Comparing with deep learning}

We also compared the simple linear models with more complex and deeper models, implemented using PyTorch. Here we considered the most direct extension of a linear model to a more complex model, adding a single hidden layer into the network. We vary the number of hidden layer features (neurons) between (1, 8, 16,  32) to determine if an advantage of a deeper model is achieved.

We trained the networks with stochastic gradient descent as the optimizer, mean square error as loss function, $L^2$-regularization of $10^-3$, using randomly sampled batches with batch-size of 256, for $10^4$ iterations. For the choice of activation function, we tested 4 options: identity, Leaky ReLU, a cubic polynomial activation function, as well as a sigmoid activation (rescaled to the range -5, +5).

The model with 1 hidden layer feature and identity activation is analogous to the linear model we evaluated in the main part of this work.

Overall the best performance obtained had almost identical performance to the linear models, and thus the simpler linear models (with ease of use and interpretability) are the better choice.

In addition to the deep models being applied to the supervoxel representations, we also performed preliminary experiments training a 3D ResNet-18 for chronological age prediction (from scratch) directly from the CTA images (with lungs, esophagus, and stomach masked out with the help of TotalSegmentator) in resampled to a fixed $1.5\mathtt{mm}^3$ resolution. The resulting models exhibited much lower performance than the supervoxel-based models, were prone to overfitting (making model selection very challenging), were computationally costly to train, and their predictions lacked interpretability. With a much larger dataset, the end-to-end training of 3D neural networks might be a viable approach and can, however, not be ruled out by these experiments.

\begin{table}[ht]
\centering
\caption{Evaluation of deeper (two-layer) models with various numbers of hidden layer features, and various activation functions after the hidden layer neurons.}
\label{tab:comparewithdeep}
\begin{tabular}{c|c|c|c|c}
Sex & Hidden Layer Features & Activation  & $\mathtt{MAE} \downarrow$ & $R^2 \uparrow$\\ \hline \hline
Female & \multirow{2}{*}{1} & \multirow{2}{*}{Identity} & 2.79 & 0.430 \\
Male &  & & 2.81 & 0.377 \\ \hline \hline
Female & \multirow{2}{*}{8} & \multirow{2}{*}{Leaky ReLU} & 3.01 & 0.346 \\
Male &  & & 3.11 & 0.275 \\ \hline 
Female & \multirow{2}{*}{16} & \multirow{2}{*}{Leaky ReLU} & 3.02 & 0.364 \\
Male &  & & 3.06 & 0.300 \\ \hline 
Female & \multirow{2}{*}{32} & \multirow{2}{*}{Leaky ReLU}  & 3.05 & 0.351 \\
Male &  & & 3.09 & 0.294 \\ \hline 
Female & \multirow{2}{*}{64} & \multirow{2}{*}{Leaky ReLU}  & 3.08 & 0.346 \\
Male &  & & 3.12 & 0.291 \\ \hline 
Female & \multirow{2}{*}{8} & \multirow{2}{*}{Cubic} & 2.80 & 0.427 \\
Male &  & & 2.82 & 0.377 \\ \hline 
Female & \multirow{2}{*}{16} & \multirow{2}{*}{Cubic} & 2.80 & 0.427 \\
Male &  & & 2.83 & 0.374 \\ \hline 
Female & \multirow{2}{*}{32} & \multirow{2}{*}{Cubic} & 2.79 & 0.429 \\
Male &  & & 2.82 & 0.376 \\ \hline 
Female & \multirow{2}{*}{64} & \multirow{2}{*}{Cubic} & 2.79 & 0.429 \\
Male &  & & 2.82 & 0.377 \\ \hline 
Female & \multirow{2}{*}{8} & \multirow{2}{*}{Sigmoid} & 2.75 & 0.436 \\
Male &  & & 2.78 & 0.382 \\ \hline 
Female & \multirow{2}{*}{16} & \multirow{2}{*}{Sigmoid} & 2.76 & 0.434 \\
Male &  & & 2.79 & 0.381 \\ \hline 
Female & \multirow{2}{*}{32} & \multirow{2}{*}{Sigmoid} & 2.75 & 0.435 \\
Male &  & & 2.79 & 0.380 \\ \hline 
Female & \multirow{2}{*}{64} & \multirow{2}{*}{Sigmoid} & 2.76 & 0.433 \\
Male &  & & 2.79 & 0.378 \\ \hline 
\end{tabular}
\end{table}

\section{Filtering based on volume prediction performance as quality control}

One possible reason for the variability in the chronological age prediction performance is noisy and distorted features introduced during the image acquisition, segmentation, registration process, or feature extraction processes. Here we investigate if filtering out all individuals who have an absolute error in volume prediction which constitutes an outlier (using 1.5 IQR filtering) on at least one of the basic volume prediction tasks (LVV, RVV, LAV, RAV, MYOV) can lead to improvements in chronological age prediction, i.e. if the overall performance we observe is due to the features corresponding to those individuals being too corrupted to even facilitate prediction of basic volume measurements.

The resulting chronological age prediction for females has the performance: $\mathtt{MAE}=2.76$, and $R^2=0.433$, and for males: $\mathtt{MAE}=2.80$, and $R^2=0.371$, which is slightly lower than what we observed on the full dataset. Therefore we can conclude that corruption of the features is not a central factor driving the overall performance.

%\vskip3pt

%\bio{}
%Author biography without author photo.
%Author biography. Author biography. Author biography.
%Author biography. Author biography. Author biography.
%Author biography. Author biography. Author biography.
%Author biography. Author biography. Author biography.
%Author biography. Author biography. Author biography.
%Author biography. Author biography. Author biography.
%Author biography. Author biography. Author biography.
%Author biography. Author biography. Author biography.
%Author biography. Author biography. Author biography.
%\endbio

%\bio{figs/pic1}
%Author biography with author photo.
%Author biography. Author biography. Author biography.
%Author biography. Author biography. Author biography.
%Author biography. Author biography. Author biography.
%Author biography. Author biography. Author biography.
%Author biography. Author biography. Author biography.
%Author biography. Author biography. Author biography.
%Author biography. Author biography. Author biography.
%Author biography. Author biography. Author biography.
%Author biography. Author biography. Author biography.
%\endbio

%\bio{figs/pic1}
%Author biography with author photo.
%Author biography. Author biography. Author biography.
%Author biography. Author biography. Author biography.
%Author biography. Author biography. Author biography.
%Author biography. Author biography. Author biography.
%\endbio

\end{document}